\pdfoutput=1
\documentclass[11pt]{article}
\usepackage[final]{acl}  
\usepackage{amsmath}
\usepackage{amssymb}
\usepackage{amsfonts}
\usepackage{times}
\usepackage{latexsym}
\usepackage[T1]{fontenc}
\usepackage[utf8]{inputenc}
\usepackage{microtype}
\usepackage{inconsolata}
\usepackage{graphicx}
\usepackage{booktabs}
\usepackage{multirow}
\usepackage{url}
\usepackage{color}
\usepackage{xcolor}
\usepackage{algorithm}
\usepackage{algorithmic}
\usepackage{colortbl}
\usepackage{listings}
\usepackage{adjustbox}
\usepackage[most]{tcolorbox} 
\usepackage[normalem]{ulem}
\useunder{\uline}{\ul}{}

\tcbset{
    promptboxstyle/.style={
        colframe=blue!75!black,    
        colback=blue!5,           
        coltitle=white,           
        fonttitle=\bfseries,      
        sharp corners,            
        width=14cm,                
        boxrule=1pt,              
        arc=4mm,                  
        drop shadow,              
        enhanced,                
    }
}

\tcbset{
    exampleboxstyle/.style={
        colframe=gray!75!black,    
        colback=blue!5,           
        coltitle=white,           
        fonttitle=\bfseries,      
        sharp corners,            
        width=14cm,               
        boxrule=1pt,              
        arc=4mm,                  
        drop shadow,             
        enhanced,                
    }
}

\title{Meta-rater: A Multi-dimensional Data Selection Method for Pre-training Language Models}

\author{Xinlin Zhuang$^{2,1}$\thanks{These authors contributed equally to this work.},
    Jiahui Peng$^{1*}$, 
    Ren Ma$^{1*}$\thanks{Project lead.}, 
    Yinfan Wang$^1$, 
    Tianyi Bai$^1$, \\
    \textbf{Xingjian Wei$^1$, 
    Jiantao Qiu$^1$, 
    Chi Zhang$^1$, 
    Ying Qian$^2$, 
    Conghui He}$^1$\thanks{Corresponding author.}  
         \\ \normalsize {$^1$ Shanghai Artificial Intelligence Laboratory}
        \\ \normalsize {$^2$ School of Computer Science and Technology, East China Normal University}
        \\ \normalsize \texttt{xinlinzhuang@stu.ecnu.edu.cn, maren@pjlab.org.cn, heconghui@pjlab.org.cn}
        }

\definecolor{darkgreen}{RGB}{0,150,0}
\newcommand{\posdiff}[1]{\textcolor{darkgreen}{\textsuperscript{$\uparrow$#1}}}
\newcommand{\negdiff}[1]{\textcolor{red}{\textsuperscript{$\downarrow$#1}}}

\begin{document}
\maketitle
\begin{abstract}
The composition of pre-training datasets for large language models (LLMs) remains largely undisclosed, hindering transparency and efforts to optimize data quality—a critical driver of model performance. Current data selection methods, such as natural language quality assessments, diversity-based filters, and classifier-based approaches, are limited by single-dimensional evaluation or redundancy-focused strategies. To address these gaps, we propose four dimensions to evaluate data quality: professionalism, readability, reasoning, and cleanliness. We further introduce \textbf{Meta-rater}, a multi-dimensional data selection method that integrates these dimensions with existing quality metrics through learned optimal weightings. Meta-rater employs proxy models to train a regression model that predicts validation loss, enabling the identification of optimal combinations of quality scores. Experiments demonstrate that Meta-rater \textbf{doubles convergence speed} for 1.3B parameter models and improves downstream task performance by \textbf{3.23\%}, with advantages that scale to models as large as 7.2B parameters. Our work establishes that holistic, multi-dimensional quality integration significantly outperforms conventional single-dimension approaches, offering a scalable paradigm for enhancing pre-training efficiency and model capability. To advance future research, we release scripts, data, and models at \url{https://github.com/opendatalab/Meta-rater}.
\end{abstract}
 
\section{Introduction}
Large language models (LLMs) demonstrate impressive performance across various tasks, with their core capabilities primarily formulated during the pre-training process~\cite{albalak2024survey}. 
However, there is a significant lack of transparency regarding the pre-training data utilized by both open-source and proprietary LLMs.
This dearth of information hinders researchers' understanding of the detailed composition of the pre-training data employed in current trending LLMs. 
Therefore, the focus of contemporary research is shifting towards enhancing the quality of pre-training data through data selection methods, which aim to extract high-quality data from original datasets \cite{albalak2024survey, dsir,qurating, mates}. 
A series of systematic pipeline methods \cite{dolma, fineweb, d4, c4} have emerged to address data processing challenges, with data selection standing out as the most crucial component for optimizing training efficiency and model performance through high-quality data curation.

\begin{figure}[!tb]
    \centering
    \includegraphics[width=0.85\linewidth]{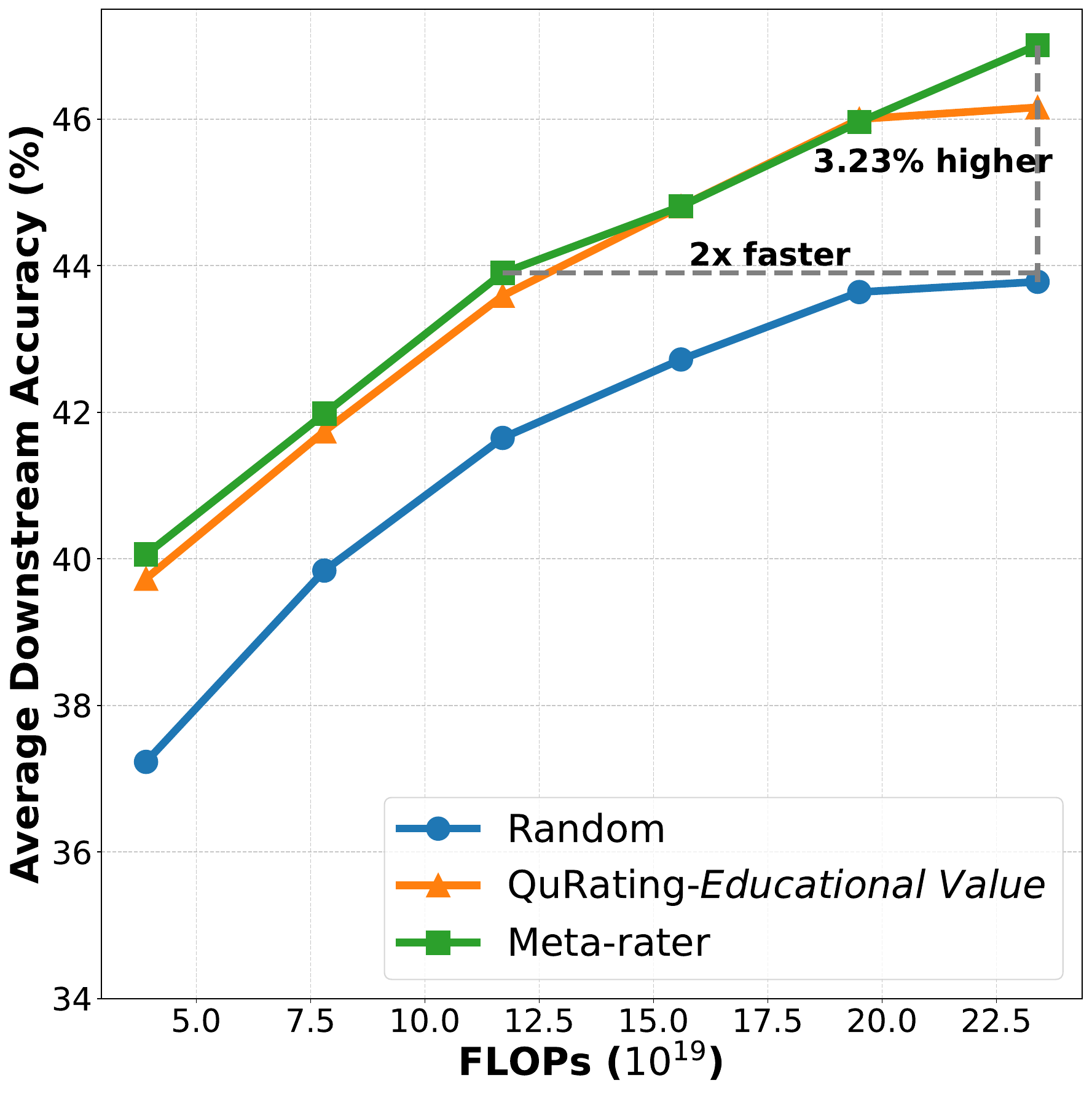}
    \caption{Comparison of average downstream task performance: random sampling, previous SOTA baseline (QuRating-\textit{Educational Value}), and our Meta-rater for pre-training a 1.3B model from scratch.}
    \label{fig:comparison}
\end{figure}

\par
Existing pre-training data selection methods can be categorized into three primary approaches: natural language quality-based methods \cite{gopher, weber2024redpajama, dsir, ppl}, diversity-based methods \cite{semdedup, softdedup, quad, bai2024multi}, and classifier-based methods \cite{qurating, fineweb}.
Alternative strategies such as MATES \cite{mates} utilize influence scores \cite{ifscore} and Rho \cite{rho} formulates data selection at the token level.
However, these methods exhibit inherent constraints - natural language quality assessment focuses on superficial text characteristics, diversity-based approaches prioritize redundancy reduction over intrinsic quality evaluation, and classifier-based techniques operate through single-dimensional quality filters. 
This raises the fundamental question: \textit{How can we systematically integrate complementary quality dimensions to achieve holistic data selection?}
\par

To address this gap, we develop four novel evaluation dimensions \textbf{PRRC} (\textbf{P}rofessionalism, \textbf{R}eadability, \textbf{R}easoning, and \textbf{C}leanliness) to expand current quality metrics. 
Utilizing these dimensions, we introduce \textbf{Meta-rater}, a model-based framework that strategically integrates multiple quality scores for optimal data selection. 
Meta-rater operates by training small proxy models and fitting a model on their data, thereby deriving optimal combination of various quality scores.
Empirical validation demonstrates Meta-rater's efficacy across model scales: for 1.3B parameter models trained on 30B tokens, it achieves \textbf{twice} the convergence speed compared to random selection and a \textbf{3.23\%} overall performance improvement. 
Scalability is evidenced with 3.3B and 7.2B models, where performance gains persist.
These results substantiate that integrating multi-dimensional quality metrics surpasses conventional single-dimension approaches, establishing a new paradigm for data curation in LLM development.

\par
Our contributions are summarized as follows:
\begin{itemize}
    \item \textbf{PRRC Framework}: We propose four novel evaluation dimensions (\textbf{P}rofessionalism, \textbf{R}eadability, \textbf{R}easoning, and \textbf{C}leanliness) to comprehensively assess pre-training data quality, supported by fine-tuned rating models that achieve \textbf{87–92\%} F1 scores, expanding beyond existing heuristic metrics.
    \item \textbf{Annotated SlimPajama-627B}: We release the first fully annotated 627B-token SlimPajama, labeled across \textbf{25 quality metrics} (including natural language features, domain importance weights, and model-based ratings), providing a foundational resource for data-centric LLM research.  
    \item \textbf{Meta-rater Methodology}: We introduce a scalable framework for \textbf{multi-dimensional} data selection, leveraging proxy models and regression analysis to derive optimal quality score weightings, advancing beyond single-dimensional filtering.
    \item \textbf{Empirical Validation}: We demonstrate Meta-rater's practical impact—\textbf{doubled convergence speed} and \textbf{3.23\%} downstream task improvement for 1.3B models—with scalability validated on 3.3B and 7.2B models.  
\end{itemize}

\section{Related Work}
As the scale of training corpora continues to grow and data-centric AI evolves, there is an increasing need for systematic approaches to select high-quality pre-training data. 
This need has spurred the development of comprehensive pre-training data processing pipelines \cite{refinedweb, he2023wanjuan, he2024opendatalab}, and data selection methods.
Existing pre-training data selection methods can be categorized into three primary approaches: natural language quality-based methods, diversity-based methods, and classifier-based methods.
\par
For natural language quality-based methods, Gopher \cite{gopher} and RedPajama \cite{weber2024redpajama} propose empirical rules like controlling the ratio of word and number tokens in texts to improve language modeling.
Additionally, previous works ~\cite{muennighoff2024scaling, wenzek2020ccnet} have shown that selecting data with perplexity (PPL) scores on validation datasets can lead to superior performance on downstream tasks compared to using the entire dataset. 
Another notable method is DSIR~\cite{dsir}, which streamlines the selection process by employing hashed N-gram features (named as data importance scores) to efficiently identify high-quality data within large datasets.
Meanwhile, another line of works utilize clustering \cite{quad} or deduplication \cite{semdedup, softdedup} to enhance diversity of pre-training datasets.
\par
More recently, more model-based classifiers have been introduced to assess the quality of pre-training data for LLMs. 
WanjuanCC \cite{qiu2024wanjuan} employs two BERT-based classifiers to filter out data containing excessive advertisements and exhibiting lower fluency.
QuRating~\cite{qurating} introduces an innovative framework that simulates human-like text quality assessments, proposing four criteria to guide data selection.
Similarly, Fineweb-Edu~\cite{fineweb} focuses specifically on assessing the \textit{Educational Value} of data.
Dataman \cite{dataman} defines 14 quality criteria and 15 domain-specific prompts, leveraging GPT-4-Turbo to evaluate document quality.
These advancements underscore the importance of optimizing data selection techniques to enhance the efficiency and effectiveness of language model training. 
However, these methods either have high computational costs or consider data quality from only a limited number of aspects. 
In contrast, Meta-rater evaluates data quality across multiple dimensions, balancing different rating criteria to achieve more effective pre-training data selection.

\section{Meta-rater}
\label{sec:dataselection}
\subsection{Task Formulation}
Data selection aims to identify the most valuable training examples from a large corpus to accelerate model learning, improve downstream task performance, and reduce computational costs.
It is formulated as selecting a subset of data $D_s$ from a large corpus $D$ to maximize the performance of a language model $\pi_\theta$ on a set of downstream tasks $T$, measured by a lower loss on validation set $J(\theta)$:
\begin{equation}
    D_s = \mathop{\arg\min}\limits_{D_s \subset D} J(\theta)
\end{equation}
where $J(\theta)$ is the loss function of the pre-trained language model $\pi_\theta$ on the validation set $V$.

Previously, this task was typically completed by top-\textit{k} selection based on a single quality score measuring one dimension. In this work, we extend this task to incorporate multiple quality scores covering different dimensions. The challenge then becomes how to aggregate these scores to derive a final data quality score:
\begin{equation}
    Q_{agg} = F(Q_1, Q_2, ..., Q_m)
\end{equation}
where $Q_{agg}$ is the final aggregated data quality score, $Q_1, Q_2, ..., Q_m$ represent various quality scores across different dimensions, and $F$ is the aggregation function that combines multiple quality scores into a single one.

\begin{algorithm}[t]
    \caption{Meta-rater}
    \label{alg:main}
    \begin{algorithmic}
    \REQUIRE Training data $\mathcal{D}$ with $m$ quality scores $\mathbf{q} = \{q_1, q_2, \dots, q_m\}$  for each example, validation dataset $\mathcal{D}_v$, number of proxy models $N$.
    \STATE \textbf{Output:} Optimal weights $\mathbf{w}^* = \{w_1^*, w_2^*, \dots, w_m^*\}$ for $m$ quality scores.
    \FOR{$i = 1, \dots, N$}
    \STATE Generate random weights $\mathbf{w_i}$ for $m$ quality scores.
    \STATE Select data from $\mathcal{D}$ based on $\mathbf{w_i}^T\mathbf{q}$, which results in $\mathcal{D}_i$.
    \STATE Train a proxy model $\mathcal{M}_i$ on the dataset $\mathcal{D}_i$.
    \STATE Compute model loss $l_i$ on validation dataset $\sum_{i=1}^{|\mathcal{D}_v|} \{\mathcal{L_{M}}_i(x_i) \mid x_i \in \mathcal{D}_v\}$.
    \ENDFOR
    \STATE Train a model $f(\mathbf{w})$ on $\{(\mathbf{w}_i, l_i)\}_{i=1}^N$ to predict $l$.
    \STATE Simulate weights $\mathbf{\tilde{w}}$ in a larger space, and predict the corresponding loss: $\hat{l} = f(\mathbf{\tilde{w}})$.
    \STATE Identify the $\mathbf{w}^*$ that minimizes $\hat{l}$: $\mathbf{w}^* = \arg\min_{\mathbf{\tilde{w}}} f(\mathbf{\tilde{w}})$.
    \STATE \textbf{Return:} Optimal quality scores weights $\mathbf{w}^*$.
    \end{algorithmic}
    \label{alg:meta_rater}
\end{algorithm}

\subsection{Meta-rater Design}

To address the aforementioned challenge, we introduce a framework called \textbf{Meta-rater}, designed to combine multiple data quality scores into a single aggregated score for data selection.

The goal of Meta-rater is to identify the optimal strategy for combining data quality scores to achieve the lowest validation loss. Essentially, Meta-rater approaches this as a regression modeling problem, fitting a model using data generated by hundreds of small-scale proxy models. 
A complete workflow of Meta-rater is provided in Algorithm \ref{alg:meta_rater}.
While inspired by \citet{regmix}, which optimizes domain mixing weights via regression, Meta-rater generalizes this approach to a broader class of data selection tasks. Data mixing, as in \citet{regmix}, becomes a special case where quality scores correspond to domain classifiers. 

\paragraph{Data Collection from Proxy Models.}
Suppose we need $N$ proxy models, then we need to do the following processes for $N$ times:
\begin{enumerate}
\item Generate a set of random weights $\mathbf{w}_i = \{w_{i1}, w_{i2}, \dots, w_{im}\}$ for $m$ quality scores, where each weight represents the importance of a specific quality dimension. 
\item Calculate the aggregated quality score for each data example $x$ as a weighted sum: $Q_{agg}(x) = \sum_{j=1}^{m} w_{ij} \cdot Q_j(x)$, where $Q_j(x)$ is the quality score of dimension $j$ for example $x$.
\item Select top-\textit{k} data examples based on their aggregated quality scores $Q_{agg}(x)$ to form a training dataset $\mathcal{D}_i$.
\item Train a small-scale proxy model $\mathcal{M}_i$ on the selected dataset $\mathcal{D}_i$ for a fixed number of steps.
\item Evaluate the proxy model $\mathcal{M}_i$ on a validation set $\mathcal{D}_v$ to obtain a validation loss $l_i = \mathcal{L}(\mathcal{M}_i, \mathcal{D}_v)$.
\end{enumerate}
After completing this process $N$ times with different sets of random weights, we obtain $N$ data points $\{(\mathbf{w}_i, l_i)\}_{i=1}^N$ that map quality score weights to validation losses.

\paragraph{Model Fitting and Optimal Weight Prediction.} 
Using the collected data points $\{(\mathbf{w}_i, l_i)\}_{i=1}^N$, we fit a regression model $f$ that predicts validation loss given a set of quality score weights: $\hat{l} = f(\mathbf{w})$. We employ a LightGBM regression model to capture non-linear relationships between quality score weights and validation loss.

With the fitted regression model, we can efficiently explore the space of possible weight combinations without requiring additional training runs. Specifically, we:
\begin{enumerate}
\item Generate a large number of candidate weight combinations $\{\mathbf{\tilde{w}}_j\}_{j=1}^J$ that cover the weight space more densely than the initial random samples.
\item Use the regression model to predict the validation loss for each candidate: $\hat{l}_j = f(\mathbf{\tilde{w}}_j)$.
\item Identify the optimal weights $\mathbf{w}^* = \arg\min_{\mathbf{\tilde{w}}_j} f(\mathbf{\tilde{w}}_j)$ that yield the minimum predicted validation loss.
\end{enumerate}

Finally, the optimal weights $\mathbf{w}^*$ are used to compute the aggregated quality scores for all data examples, and the top-ranked examples are selected for training the final language model. To enhance robustness, we average the top-$k$ predicted weight combinations rather than using only the single best prediction.

\subsection{Data Quality Scores}
\label{sec: PRRC}

To provide a comprehensive evaluation of data quality, we employ a multi-faceted approach that combines natural language quality signals, data importance weights, and model-based heuristic ratings. These methods collectively enable us to assess the linguistic integrity, domain relevance, and semantic depth of textual data. The following subsections detail each of these components, outlining the specific metrics and methodologies used to ensure a robust and thorough analysis of data quality.
A full list of all quality scores and corresponding explanations is provided in Appendix \ref{app:redpajama}.

\paragraph{Natural Language Quality Signals.}
We choose rule-based measures proposed by RedPajama \cite{weber2024redpajama} indicating how natural a given piece of text is, including the number of sentences and words, the fraction of non-alphabet words, etc.

\paragraph{Data Importance Scores.}
Data importance scores measure how similar a given text is to a high-quality domain based on hashed N-gram features \cite{dsir}. 
In addition to \textit{Book} and \textit{Wikipedia}, we also consider the importance weights compared to \textit{AutoMathText} \cite{automathtext} to account for the math domain.

\paragraph{Model-based Ratings.}

Recent studies have employed classifiers to filter data based on human-defined heuristic criteria, such as educational value \cite{fineweb}, fluency \cite{qiu2024wanjuan}, and writing style \cite{qurating}. These classifiers, typically built on learnable transformer models like fine-tuned BERT, are capable of capturing deep semantic features of text.
We utilized the \textit{Advertisement} and \textit{Fluency} dimensions from WanjuanCC \cite{qiu2024wanjuan}, the \textit{Educational Value} dimension from Fineweb-edu \cite{fineweb}, and the four dimensions from QuRating \cite{qurating}. 
Building on existing research on data quality and our practical insights, we further introduce four additional dimensions PRRC to ensure a more comprehensive assessment of data quality.

\begin{enumerate}
    \item \textit{Professionalism}. This dimension serves as an indicator of the level of professional knowledge contained in the text. LLMs trained with sufficient professional corpus (e.g., textbooks, research articles) demonstrate superior performance in examinations and general QA tasks \cite{gunasekar2023textbooks}. Building on the \textit{Required Expertise} used in QuRating \cite{qurating}, we have refined this dimension by developing a more detailed rating scale and implementing more precise rating criteria. 
    
    \item \textit{Readability}. In linguistics, readability refers to the ease with which a reader can understand a written text \cite{dubay2004principles}. We believe that readability is equally crucial for LLM pre-training. Educators have developed several formulas to assess readability, which typically consider factors such as sentence length, word length, syllable count, and word frequency. 
    
    \item \textit{Reasoning}. With the introduction of OpenAI's o1 model, LLMs have transitioned into the era of reasoning models. Research by DeepSeek has shown that smaller language models can achieve reasoning capabilities on par with LLMs that are ten times their size by leveraging supervised fine-tuning on high-reasoning data \cite{guo2025deepseek}. This finding highlights the critical importance of data rich in reasoning complexity. To address this, we developed this dimension to identify data exhibiting exceptional reasoning depth. Such data typically involves multi-step logical reasoning, thorough analysis, and requires readers to synthesize diverse information to form well-rounded conclusions.
     
     \item \textit{Cleanliness}. A clean text should be formatted correctly as complete sentences, without inappropriate characters, with an appropriate length and minimal noise (e.g., hyperlinks, advertisements, irrelevant information). Prior research has demonstrated the substantial benefit of clean data for LLM pre-training \cite{fineweb}. 
     In contrast to the other three dimensions that focus on semantic features, this dimension aims to capture the literal characteristics of given texts. 
     We consolidate relative criteria into a single dimension fitted by a model instead of using heuristic rules because model-based approach exhibits superior generalization capability in dealing with irregular, long-tailed anomalies present in text.
    
\end{enumerate}

To quantify the quality of pre-training data along aforementioned four dimensions, we implement an additive 5-point rating system in which points are awarded incrementally based on meeting specific criteria. 
For each dimension, we have developed corresponding prompt and rating model.
Specifically, we employ Llama-3.3-70B-Instruct\footnote{\url{https://huggingface.co/meta-llama/Meta-Llama-3.3-70B-Instruct}} to rate the quality of 500k examples sampled from SlimPajama, which thereby constitutes the training data for our quality rating models.
With these data, we fine-tune ModernBERT \cite{modernbert} as the rating model for each dimension.
These models achieve F1 scores of 91.57\% for \textit{Professionalism}, 87.47\% for \textit{Readability}, 89.59\% for \textit{Reasoning}, and 87.88\% for \textit{Cleanliness} on the test set. 
Further details regarding prompts for rating data, rating models, and training are provided in Appendix \ref{app:prrc-models}.

\section{Experiment}
\subsection{Experimental Setup}

\begin{table*}[!t]
\centering
\small
\scalebox{0.85}{
\resizebox{\linewidth}{!}{
\begin{tabular}{@{}lcccc@{}}
\toprule
\textbf{Data Selection Method} &
  \textbf{\begin{tabular}[c]{@{}c@{}}General\\ Knowledge\end{tabular}} &
  \textbf{\begin{tabular}[c]{@{}c@{}}Commonsense\\ Reasoning\end{tabular}} &
  \textbf{\begin{tabular}[c]{@{}c@{}}Reading\\ Comprehension\end{tabular}} &
  \textbf{Average} \\ \midrule
Random (30B Tokens)& 52.79 \phantom{\posdiff{0.00}} & 43.94 \phantom{\posdiff{0.00}} & 30.02 \phantom{\posdiff{0.00}} & 43.78 \phantom{\posdiff{0.00}} \\
Random (60B Tokens)& 56.01 \posdiff{3.22} & 44.87 \posdiff{0.93} & 31.47 \posdiff{1.45} & 45.70 \posdiff{1.92} \\ \midrule
PPL                                & 52.53 \negdiff{0.26} & 40.53 \negdiff{3.41} & 26.52 \negdiff{3.50} & 41.53 \negdiff{2.25} \\ \midrule
Semdedup                           & 52.65 \negdiff{0.14} & 42.66 \negdiff{1.28} & 28.92 \negdiff{1.10} & 42.97 \negdiff{0.81} \\ \midrule
DSIR                               &       &       &                      &       \\
\quad Target as \textit{Book}      & 52.45 \negdiff{0.34} & \textbf{46.93} \textbf{\posdiff{2.99}} & 28.94 \negdiff{1.08} & 44.50 \posdiff{0.72} \\
\quad Target as \textit{Wikipedia} & 54.94 \posdiff{2.15} & 41.87 \negdiff{2.07} & 27.39 \negdiff{2.63} & 43.15 \negdiff{0.63} \\ \midrule
QuRating                           &       &       &                      &       \\
\quad \textit{Required Expertise}  & 56.91 \posdiff{4.12} & 45.16 \posdiff{1.22} & 28.06 \negdiff{1.96} & 45.29 \posdiff{1.51} \\
\quad \textit{Writing Style}       & 57.14 \posdiff{4.35} & 46.28 \posdiff{2.34} & 28.19 \negdiff{1.83} & 45.83 \posdiff{2.05} \\
\quad \textit{Facts and Trivia}    & 57.58 \posdiff{4.79} & 45.62 \posdiff{1.68} & 29.40 \negdiff{0.62} & 46.05 \posdiff{2.27} \\
\quad \textit{Educational Value}   & \underline{57.66} \posdiff{4.87} & 46.72 \posdiff{2.78} & 28.10 \negdiff{1.92} & 46.16 \posdiff{2.38} \\ \midrule
Fineweb-Edu                        & 55.79 \posdiff{3.00} & 45.51 \posdiff{1.57} & 31.10 \posdiff{1.08} & 45.76 \posdiff{1.98} \\ \midrule
MATES                              & 53.15 \posdiff{0.36} & 43.25 \negdiff{0.69} & 30.55 \posdiff{0.53} & 43.79 \posdiff{0.01} \\ \midrule
\textbf{PRRC (Ours)}               &       &       &                      &       \\
\quad \textit{Professionalism}     & 56.11 \posdiff{3.32} & 44.66 \posdiff{0.72} & 29.89 \negdiff{0.13} & 45.26 \posdiff{1.48} \\
\quad \textit{Readability}         & 56.18 \posdiff{3.39} & 45.41 \posdiff{1.47} & 31.20 \posdiff{1.18} & 45.89 \posdiff{2.11} \\
\quad \textit{Reasoning}           & 55.57 \posdiff{2.78} & 44.86 \posdiff{0.92} & 30.48 \posdiff{0.46} & 45.28 \posdiff{1.50} \\
\quad \textit{Cleanliness}         & 56.45 \posdiff{3.66} & 44.88 \posdiff{0.94} & 30.72 \posdiff{0.70} & 45.68 \posdiff{1.90} \\ \midrule
\textbf{Meta-rater (Ours)}         &       &       &                      &       \\
\quad PRRC (4) &
  57.01 \posdiff{4.22} &
  \underline{45.86} \posdiff{1.92} &
  31.11 \posdiff{1.09} &
  46.35 \posdiff{2.57} \\
\quad Model (11) &
  57.34 \posdiff{4.55} &
  45.62 \posdiff{1.68} &
  \textbf{31.96} \textbf{\posdiff{1.94}} &
  \underline{46.60} \posdiff{2.82} \\
\quad All (25) &
  \textbf{58.90} \textbf{\posdiff{6.11}} &
  45.41 \posdiff{1.47} &
  \underline{31.55} \posdiff{1.53} &
  \textbf{47.01} \textbf{\posdiff{3.23}} \\ \bottomrule
\end{tabular}
}
}
\caption{Performance of data selection methods on downstream tasks. For Meta-rater, the number in parentheses ( ) indicates the number of quality scores used. We report performance improvements compared to random sampling of 30B tokens, with the \textbf{best result} highlighted and the \underline{second best result} underlined in each column. \textit{Model} refers to model-based ratings, while \textit{All} denotes the inclusion of all 25 quality scores. Full evaluation results are provided in Appendix \ref{app:results}.}

\label{tab:main}
\end{table*}

\paragraph{Training.} 
We utilize SlimPajama \cite{slimpajama} as the data pool for the training set. 
For each data selection method, we sample a total of 30B tokens while maintaining a fixed domain proportion (see Appendix \ref{app:domain_weights}). 
Using each sampled dataset of 30B tokens, we train a transformer-based, decoder-only language model from scratch.
In our main experiments, we employ a model with 1.3B parameters, incorporating Rotary Positional Embeddings (RoPE) \cite{su2024roformer} and a maximum context window of 1,024 tokens. 
To further validate the effectiveness of Meta-rater, we conduct additional experiments with 3.3B-parameter language models trained on 100B tokens.
Details regarding the training process are provided in Appendix \ref{app:training}.

\paragraph{Evaluation.}
To comprehensively assess the capabilities of pre-trained models, we conduct holistic evaluations on various downstream tasks covering three significant categories: \textbf{General Knowledge} (including ARC-Challenge \cite{arc}, ARC-Easy, and SciQ \cite{sciq}), \textbf{Commonsense Reasoning} (including HellaSwag \cite{hellaswag}, SIQA \cite{siqa}, and WinoGrande \cite{winogrande}), and \textbf{Reading Comprehension} (including RACE \cite{race} and OpenbookQA \cite{openbookqa}).
Evaluations are conducted using the \textbf{lm-evaluation-harness} \cite{eval-harness} framework with in-context learning setting, and average accuracy is reported for convenient comparison. 
Further details of evaluation are shown in Appendix \ref{app:icl}.

\paragraph{Baselines.}
We compare Meta-rater with the following data selection methods:
\begin{enumerate}
\item \textbf{Random}: This method involves randomly selecting a subset from SlimPajama without applying any data quality controls.
\item \textbf{PPL} \cite{ppl}:  This approach selects a subset of samples with the lowest perplexity scores on the validation dataset.
\item \textbf{Semdedup} \cite{semdedup}: The entire SlimPajama is clustered into 10,000 clusters, and data points farthest from the centroid in each cluster are selected.
\item \textbf{DSIR} \cite{dsir}: This method employs hashed N-gram features to identify and select data that exhibits similarity to a specified dataset. We set \textit{Book} and \textit{Wikipedia} as target domains.
\item \textbf{QuRating} \cite{qurating}: We employ four quality raters from QuRating, namely \textit{Required Expertise}, \textit{Writing Style}, \textit{Facts and Trivia}, and \textit{Educational Value} for selection.
\item \textbf{Fineweb-edu} \cite{fineweb}: Similar to QuRating, we utilize the educational value rater and select top-$k$ data. 
\item \textbf{MATES} \cite{mates}: We train ModernBERT as the influence score predictor \cite{ifscore} and select a subset of samples with top-$k$ influence scores.
\item \textbf{PRRC} We use the rating models trained in Section \ref{sec: PRRC} for \textit{Professionalism}, \textit{Readability}, \textit{Reasoning}, and \textit{Cleanliness} to select data. 
\end{enumerate}

\subsection{Results and Analysis}
\subsubsection{Analysis of Quality Metrics}
\paragraph{Analysis of quality score weight distribution.}
Table \ref{tab:meta_rater_weights} presents the learned weights of all 25 quality scores, revealing significant patterns in how different quality dimensions contribute to model performance.
Our findings show that \textit{Educational Value} emerges as the most influential metric (5.64\%), confirming observations from QuRating and FineWeb-Edu research. In contrast, \textit{Writing Style} has minimal impact (0.05\%), which aligns with QuRating's observation that high Writing Style content failed to outperform random sampling. Despite their simplicity, natural language signals prove valuable, especially those that identify non-alphabetical content. Among our PRRC metrics, \textit{Reasoning} (4.44\%) and \textit{Professionalism} (4.05\%) make substantial contributions, while \textit{Cleanliness} shows comparatively lower influence (1.17\%).
These weight distributions demonstrate Meta-rater's effectiveness in identifying and appropriately weighting quality dimensions according to their genuine impact on downstream performance.

\paragraph{Correlations between quality metrics.}
We also analyzed the relationships between quality metrics by calculating Spearman correlation coefficients across all 25 quality scores using 200k examples from SlimPajama, with results visualized in Figure \ref{fig:correlation_heatmap}.
Our analysis reveals three key patterns. First, model-based metrics (including our PRRC) exhibit relatively weak correlation (<0.6) with most existing metrics, indicating they capture distinct aspects of data quality. Second, Natural Language Quality Signals demonstrate strong inter-correlation among features like word count, entropy, and sentence count (>0.85). Third, Data Importance Scores (DSIR) show remarkably high correlation with each other (>0.95) while maintaining low correlation with model-based ratings.
These observations highlight that our PRRC metrics and other model-based ratings contribute novel information beyond what traditional statistical features capture, supporting their integration into our comprehensive quality assessment framework.

\subsubsection{Results of Pre-trained Models}
\paragraph{Meta-rater outperforms all baseline models.}
We evaluate all baseline models and those trained using Meta-rater. 
Evaluation results are presented in Table \ref{tab:main}. 
Meta-rater achieves the highest performance compared to previous data selection methods. 
Specifically, it surpasses the Random-30B by a margin of 3.23 in average accuracy and exceeds the previous SOTA method, QuRating \textit{Educational Value}, by 0.85. 
Notably, Meta-rater excels across all task categories, highlighting its robustness and versatility in addressing a wide range of downstream tasks.
Additionally, we conduct evaluations on knowledge-intensive benchmarks such as MMLU \cite{mmlu} and NaturalQuestions \cite{nq}. The results, detailed in Appendix \ref{app:MMLU_NQ_results}, align with our primary findings and further confirm Meta-rater's effectiveness compared to all baseline methods.

\begin{table}
\centering
\huge
\renewcommand{\arraystretch}{1.1}
\resizebox{\linewidth}{!}{
\begin{tabular}{@{}lc@{}}
\toprule
\textbf{Process}                          & \textbf{FLOPs ($10^{19}$)} \\ \midrule
Quality Scores Rating                     & \multicolumn{1}{l}{}       \\
\quad Fineweb-edu Classifier              & 0.44                       \\
\quad WanjuanCC Classifiers (2)           & 0.88                       \\
\quad QuRating Classifiers (4)            & 6.18                       \\
\quad PRRC Classifiers (4)                & 25.52                      \\ \midrule
Meta-rater Construction                   & \multicolumn{1}{l}{}       \\
\quad Proxy Models Training and Inference & 0.18                       \\ \midrule
Pre-training                              & \multicolumn{1}{l}{}       \\
\quad 1.3B Model on 30B Tokens            & 23.40                      \\
\quad 3.3B Model on 100B Tokens           & 198.00                     \\ \bottomrule
\end{tabular}
}
\caption{FLOPs for quality scores rating, Meta-rater construction, and language model pre-training.}
\label{tab:cost}
\end{table}

\paragraph{Meta-rater is computationally efficient.}
As shown in Figure \ref{fig:comparison}, Meta-rater matches the performance of Random-30B model using only 15B tokens. 
When consuming 30B tokens, it surpasses the Random-60B model by a margin of 1.31, despite using half the number of tokens.
To quantify computational efficiency, we analyze the FLOPs required for quality score rating, Meta-rater construction, and language model pre-training, with detailed breakdowns provided in Appendix \ref{app:cost}.
As shown in Table \ref{tab:cost}, the FLOPs for Meta-rater constitute only 0.7\% of those required to pre-train a 1.3B model. Although the FLOPs for quality score rating are approximately 1.4 times higher than pre-training a 1.3B model, the annotated labels generated are reusable for various purposes and represent a valuable resource for the broader research community.
Furthermore, the cost-effectiveness of the rating process becomes increasingly pronounced at larger pre-training scales: it accounts for only 17\% of the FLOPs required to pre-train a 3.3B model on 100B tokens. In summary, Meta-rater demonstrates significant advantages in enabling efficient and scalable pre-training.

\paragraph{Scalability on the number of quality scores.}
We conduct experiments to investigate the impact of the number of quality scores on Meta-rater's performance. In addition to the default setting of 25 quality scores, we evaluate models trained using only PRRC ratings (4 quality scores) and a combination of all model-based ratings (WanjuanCC + QuRating + FineWeb-Edu + PRRC = 11 quality scores).
As shown in Table \ref{tab:main}, performance improves progressively as the number of quality scores increases: 46.35 (4) → 46.60 (11) → 47.01 (25). This trend suggests that Meta-rater continues to benefit from incorporating additional raters, highlighting the potential for further gains with an expanded set of quality metrics.

\paragraph{Scaling to larger models and datasets.}
We conduct additional experiments to evaluate Meta-rater's effectiveness when scaling to larger models and datasets. Specifically, we pre-train 3.3B models using 100B tokens and 7.2B model using 150B tokens from scratch, comparing random sampling against Meta-rater with all 25 raters.
As shown in Table \ref{tab:scaling}, Meta-rater consistently outperforms random sampling across all model sizes and training data amounts. For the 3.3B model trained on 100B tokens, Meta-rater achieves an average score of 54.71, surpassing the random sampling baseline of 52.98 by 1.73. The improvement is particularly pronounced in General Knowledge tasks (67.51 vs 64.22).
For the 7.2B model trained on 150B tokens, the gap widens further, with Meta-rater outperforming random sampling by 3.12 (55.24 vs 52.12).
These results demonstrate that Meta-rater's benefits scale effectively to larger models and datasets, consistently delivering more efficient training across different model capacities.

\begin{table}[!tb]
\centering
\renewcommand{\arraystretch}{1.1}
\resizebox{\linewidth}{!}{
\begin{tabular}{@{}clcccc@{}}
\toprule
\multicolumn{1}{l}{\textbf{Model}} &
  \textbf{Method} &
  \multicolumn{1}{l}{\textbf{G.K.}} &
  \multicolumn{1}{l}{\textbf{C.R.}} &
  \multicolumn{1}{l}{\textbf{R.C.}} &
  \multicolumn{1}{l}{\textbf{Avg.}} \\ \midrule
\multirow{2}{*}{3.3B} & Random        & 64.22 & 53.55 & 35.28 & 52.98 \\
                      & Meta-rater    & 67.51 & 54.35 & 36.06 & 54.71 \\ \midrule
\multirow{2}{*}{7.2B} & Random        & 65.10 & 52.01 & 35.87 & 52.12 \\
                      & Meta-rater    & 67.97 & 54.58 & 37.14 & 55.24 \\ \bottomrule
\end{tabular}
}
\caption{Performance of 3.3B and 7.2B models with random sampling and Meta-rater on downstream tasks. Abbreviations: G.K. = General Knowledge, C.R. = Commonsense Reasoning, R.C. = Reading Comprehension.}
\label{tab:scaling}
\end{table}

\section{Analysis}

\begin{table}
\centering
\resizebox{\linewidth}{!}{
\begin{tabular}{@{}lcccc@{}}
\toprule
\textbf{Method}            & \textbf{G.K.}   & \textbf{C.R.}   & \textbf{R.C.}   & \textbf{Avg.}   \\ \midrule
Random                     & 52.79           & 43.94           & 30.02           & 43.78           \\ \midrule
Mean                       &                 &                 &                 &                 \\
\quad PRRC (4)             & 55.04           & 42.42           & 30.34           & 44.13           \\
\quad Model (11)           & \underline{58.05} & 42.38           & 31.30           & 45.49           \\
\quad All (25)             & 56.29           & 42.30           & 30.74           & 44.65           \\ \midrule
Intersection               &                 &                 &                 &                 \\
\quad QuRating (4)         & 54.57           & 42.74           & 31.30           & 44.31           \\
\quad PRRC (4)             & 55.83           & 44.05           & 30.86           & 45.17           \\ \midrule
\textbf{Meta-rater (Ours)} &                 &                 &                 &                 \\
\quad PRRC (4)             & 57.01           & \textbf{45.86}  & 31.11           & 46.35           \\
\quad Model (11)           & 57.34           & \underline{45.62} & \textbf{31.96}  & \underline{46.60} \\
\quad All (25)             & \textbf{58.90}  & 45.41           & \underline{31.55} & \textbf{47.01}  \\ \bottomrule
\end{tabular}

}
\caption{Downstream task results comparison of naive rater combination methods.}
\label{tab:combination}
\end{table}

\subsection{Effect of Proxy Models}
\begin{figure}[!tb]
    \centering
    \includegraphics[width=1.0\linewidth]{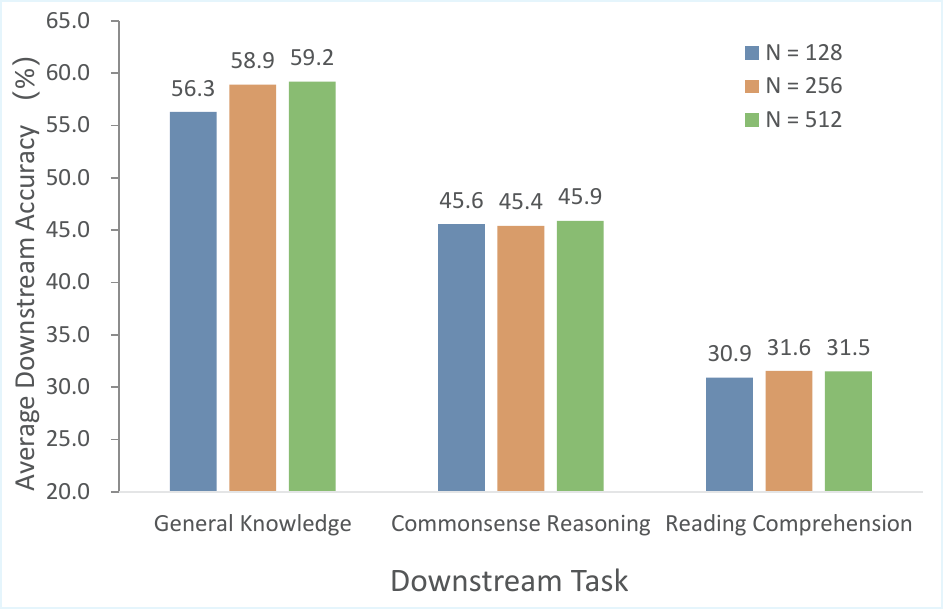}
    \caption{Average downstream task performance of Meta-rater with different numbers of proxy models. }
    \label{fig:regression}
\end{figure}

\paragraph{Number.}
We analyze the impact of the number of proxy models (\textit{N}) on Meta-rater's performance, with results illustrated in Figure \ref{fig:regression}. The overall trend reveals that increasing \textit{N} leads to significant performance improvements, particularly in General Knowledge tasks, which exhibit the most substantial gains compared to other task categories. However, the marginal improvements diminish as \textit{N} increases from 256 to 512. Based on these observations, we identify $N$=256 as an optimal choice, striking a balance between performance gains and efficiency.

\paragraph{Proxy Model Architecture.}
Table \ref{tab:architecture} provides details on our original proxy model architecture. To examine if proxy model architecture affects data selection outcomes, we expand the model size from 18M to 46M parameters by increasing hidden dimensions (256$\to$512) and layer count (2$\to$4), then generate a new set of data score weights.
Comparing the datasets selected using these new weights with those from the original proxy model revealed a 94.6\% overlap between them.
This substantial agreement indicates that Meta-rater's data selection recommendations remain consistent despite moderate changes to the proxy model size.

\subsection{Effect of Combining Strategies for Quality Scores}
We explore two straightforward methods for combining quality scores as alternatives to the Meta-rater: \textit{Mean}: This approach uses the arithmetic mean of all quality scores, giving equal weight to each score, and \textit{Intersection}\footnote{Due to strict selection criteria, sufficient data for pre-training could only be obtained with 4 QuRating raters and 4 PRRC raters.}: This method selects data that meet the criteria for all quality scores.
While both methods provide ways to combine quality scores, they result in only slight performance improvements. The proposed Meta-rater outperforms the \textit{Mean} method, achieving an average score of 47.01 compared to 44.65 with 25 raters. A similar gap is observed between Meta-rater and the \textit{Intersection} method with PRRC raters (46.35 vs 45.17).

We believe this superiority stems from key limitations of these simple combinations. The \textit{Mean} approach assumes equal importance among all raters, which leads to suboptimal results when there are imbalances in scoring. As shown in Appendix \ref{app:distribution}, individual quality score distributions vary significantly, making uniform weighting ineffective. The \textit{Intersection} method results in excessive data elimination due to strict filtering criteria, where a single low score can exclude data even if other raters provide high scores.
In contrast, Meta-rater's weighted aggregation dynamically adjusts rater contributions while preserving data integrity.
The complexity of optimal weight calibration becomes evident when examining the 25-dimensional weight space. We performed PCA to visualize the loss surface in Figure \ref{fig:loss_landscape}, which reveals multiple local minima. This explains why simple approaches like uniformly weighting scores underperform compared to Meta-rater's learned weights, as the optimal region forms a relatively small "valley" in the weight space.

\begin{figure}[!tb]
    \centering
    \includegraphics[width=1.0\linewidth]{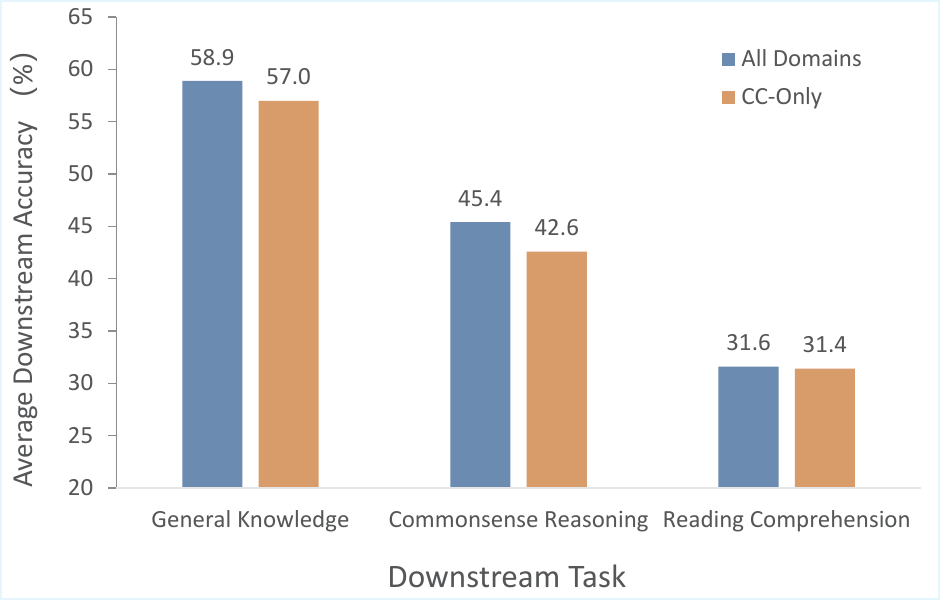}
    \caption{Average downstream task performance of Meta-rater with different settings of data domains. Abbreviation: CC = Common Crawl.}
    \label{fig:domain}
\end{figure}

\subsection{Effect of Data Domain}
Prior studies on data mixing have explored improving LLM performance on downstream tasks by adjusting the domain distribution of pre-training data. These approaches range from rule-based heuristics \cite{dmlaw, unimax} to model-driven methods \cite{doremi, doge, regmix}. 
To isolate the impact of domain diversity, we constrain data selection and pre-training to a single domain—Common Crawl—while avoiding explicit control over domain sampling ratios.
As shown in Figure \ref{fig:domain}, restricting pre-training to Common Crawl leads to a performance decline across all three task categories, with the most pronounced drop observed in Commonsense Reasoning tasks. These findings underscore the importance of domain diversity in pre-training data, highlighting that data quality alone is insufficient for maintaining robust model performance.

\section{Conclusion}
We present \textbf{Meta-rater}, a multi-dimensional framework that integrates quality metrics to identify optimal pre-training data for LLMs. 
Our evaluations demonstrate that Meta-rater consistently outperforms existing data selection methods across downstream tasks, with benefits that scale to larger models up to 7.2B parameters. 
These results confirm that Meta-rater successfully improves both efficiency and performance in LLM pre-training.

\newpage
\section*{Limitations}  
While Meta-rater demonstrates significant improvements in data selection for LLM pre-training, our study has certain limitations. 
Due to computational constraints, our experiments were conducted on relatively small-scale models (up to 7.2B parameters) and limited token budgets (150B tokens). 
Additionally, our utilized quality metrics, while comprehensive, may not fully capture all aspects of pre-training data, and we will explore refining or expanding these dimensions in the future work. 

\section*{Acknowledgement}
This work is supported by Shanghai Artificial Intelligence Laboratory. We express sincere thanks to InternTrain Team of Shanghai Artificial Intelligence Laboratory, especially Yang Gao, for their kind help for the pre-training experiments.

\bibliography{custom}

\begin{thebibliography}{49}
\expandafter\ifx\csname natexlab\endcsname\relax\def\natexlab#1{#1}\fi

\bibitem[{Abbas et~al.(2023)Abbas, Tirumala, Simig, Ganguli, and Morcos}]{semdedup}
Amro Kamal~Mohamed Abbas, Kushal Tirumala, Daniel Simig, Surya Ganguli, and Ari~S. Morcos. 2023.
\newblock \href {https://openreview.net/forum?id=4vlGm9gv6c} {Semdedup: Data-efficient learning at web-scale through semantic deduplication}.
\newblock In \emph{ICLR 2023 Workshop on Mathematical and Empirical Understanding of Foundation Models}.

\bibitem[{Albalak et~al.(2024)Albalak, Elazar, Xie, Longpre, Lambert, Wang, Muennighoff, Hou, Pan, Jeong et~al.}]{albalak2024survey}
Alon Albalak, Yanai Elazar, Sang~Michael Xie, Shayne Longpre, Nathan Lambert, Xinyi Wang, Niklas Muennighoff, Bairu Hou, Liangming Pan, Haewon Jeong, et~al. 2024.
\newblock A survey on data selection for language models.
\newblock \emph{arXiv preprint arXiv:2402.16827}.

\bibitem[{Ankner et~al.(2024)Ankner, Blakeney, Sreenivasan, Marion, Leavitt, and Paul}]{ppl}
Zachary Ankner, Cody Blakeney, Kartik Sreenivasan, Max Marion, Matthew~L Leavitt, and Mansheej Paul. 2024.
\newblock \href {https://openreview.net/forum?id=0r0Bg1NY1X} {Perplexed by perplexity: Perplexity-based pruning with small reference models}.
\newblock In \emph{ICLR 2024 Workshop on Mathematical and Empirical Understanding of Foundation Models}.

\bibitem[{Bai et~al.(2024)Bai, Yang, Wong, Peng, Zhuang, Zhang, Wu, Qiu, Zhang, Yuan et~al.}]{bai2024multi}
Tianyi Bai, Ling Yang, Zhen~Hao Wong, Jiahui Peng, Xinlin Zhuang, Chi Zhang, Lijun Wu, Jiantao Qiu, Wentao Zhang, Binhang Yuan, et~al. 2024.
\newblock Multi-agent collaborative data selection for efficient llm pretraining.
\newblock \emph{arXiv preprint arXiv:2410.08102}.

\bibitem[{Chung et~al.(2023)Chung, Garcia, Roberts, Tay, Firat, Narang, and Constant}]{unimax}
Hyung~Won Chung, Xavier Garcia, Adam Roberts, Yi~Tay, Orhan Firat, Sharan Narang, and Noah Constant. 2023.
\newblock \href {https://openreview.net/forum?id=kXwdL1cWOAi} {Unimax: Fairer and more effective language sampling for large-scale multilingual pretraining}.
\newblock In \emph{The Eleventh International Conference on Learning Representations}.

\bibitem[{Clark et~al.(2018)Clark, Cowhey, Etzioni, Khot, Sabharwal, Schoenick, and Tafjord}]{arc}
Peter Clark, Isaac Cowhey, Oren Etzioni, Tushar Khot, Ashish Sabharwal, Carissa Schoenick, and Oyvind Tafjord. 2018.
\newblock Think you have solved question answering? try arc, the ai2 reasoning challenge.
\newblock \emph{arXiv preprint arXiv:1803.05457}.

\bibitem[{Dao(2024)}]{flashattention2}
Tri Dao. 2024.
\newblock \href {https://openreview.net/forum?id=mZn2Xyh9Ec} {Flashattention-2: Faster attention with better parallelism and work partitioning}.
\newblock In \emph{The Twelfth International Conference on Learning Representations}.

\bibitem[{Devlin et~al.(2019)Devlin, Chang, Lee, and Toutanova}]{bert}
Jacob Devlin, Ming-Wei Chang, Kenton Lee, and Kristina Toutanova. 2019.
\newblock \href {https://doi.org/10.18653/v1/N19-1423} {{BERT}: Pre-training of deep bidirectional transformers for language understanding}.
\newblock In \emph{Proceedings of the 2019 Conference of the North {A}merican Chapter of the Association for Computational Linguistics: Human Language Technologies, Volume 1 (Long and Short Papers)}, pages 4171--4186, Minneapolis, Minnesota. Association for Computational Linguistics.

\bibitem[{DuBay(2004)}]{dubay2004principles}
William~H DuBay. 2004.
\newblock The principles of readability.
\newblock \emph{Online submission}.

\bibitem[{Fan et~al.(2024)Fan, Pagliardini, and Jaggi}]{doge}
Simin Fan, Matteo Pagliardini, and Martin Jaggi. 2024.
\newblock \href {https://openreview.net/forum?id=7rfZ6bMZq4} {{DOGE}: Domain reweighting with generalization estimation}.
\newblock In \emph{Forty-first International Conference on Machine Learning}.

\bibitem[{Gao et~al.(2023)Gao, Tow, Abbasi, Biderman, Black, DiPofi, Foster, Golding, Hsu, Le~Noac'h, Li, McDonell, Muennighoff, Ociepa, Phang, Reynolds, Schoelkopf, Skowron, Sutawika, Tang, Thite, Wang, Wang, and Zou}]{eval-harness}
Leo Gao, Jonathan Tow, Baber Abbasi, Stella Biderman, Sid Black, Anthony DiPofi, Charles Foster, Laurence Golding, Jeffrey Hsu, Alain Le~Noac'h, Haonan Li, Kyle McDonell, Niklas Muennighoff, Chris Ociepa, Jason Phang, Laria Reynolds, Hailey Schoelkopf, Aviya Skowron, Lintang Sutawika, Eric Tang, Anish Thite, Ben Wang, Kevin Wang, and Andy Zou. 2023.
\newblock \href {https://doi.org/10.5281/zenodo.10256836} {A framework for few-shot language model evaluation}.

\bibitem[{Gunasekar et~al.(2023)Gunasekar, Zhang, Aneja, Mendes, Del~Giorno, Gopi, Javaheripi, Kauffmann, de~Rosa, Saarikivi et~al.}]{gunasekar2023textbooks}
Suriya Gunasekar, Yi~Zhang, Jyoti Aneja, Caio C{\'e}sar~Teodoro Mendes, Allie Del~Giorno, Sivakanth Gopi, Mojan Javaheripi, Piero Kauffmann, Gustavo de~Rosa, Olli Saarikivi, et~al. 2023.
\newblock Textbooks are all you need.
\newblock \emph{arXiv preprint arXiv:2306.11644}.

\bibitem[{Guo et~al.(2025)Guo, Yang, Zhang, Song, Zhang, Xu, Zhu, Ma, Wang, Bi et~al.}]{guo2025deepseek}
Daya Guo, Dejian Yang, Haowei Zhang, Junxiao Song, Ruoyu Zhang, Runxin Xu, Qihao Zhu, Shirong Ma, Peiyi Wang, Xiao Bi, et~al. 2025.
\newblock Deepseek-r1: Incentivizing reasoning capability in llms via reinforcement learning.
\newblock \emph{arXiv preprint arXiv:2501.12948}.

\bibitem[{He et~al.(2023)He, Jin, Xu, Qiu, Wang, Li, Yan, Wang, and Lin}]{he2023wanjuan}
Conghui He, Zhenjiang Jin, Chao Xu, Jiantao Qiu, Bin Wang, Wei Li, Hang Yan, Jiaqi Wang, and Dahua Lin. 2023.
\newblock Wanjuan: A comprehensive multimodal dataset for advancing english and chinese large models.
\newblock \emph{arXiv preprint arXiv:2308.10755}.

\bibitem[{He et~al.(2024{\natexlab{a}})He, Li, Jin, Xu, Wang, and Lin}]{he2024opendatalab}
Conghui He, Wei Li, Zhenjiang Jin, Chao Xu, Bin Wang, and Dahua Lin. 2024{\natexlab{a}}.
\newblock Opendatalab: Empowering general artificial intelligence with open datasets.
\newblock \emph{arXiv preprint arXiv:2407.13773}.

\bibitem[{He et~al.(2024{\natexlab{b}})He, Xiong, Liu, Liao, Ding, Zhang, Tang, Han, and Wei}]{softdedup}
Nan He, Weichen Xiong, Hanwen Liu, Yi~Liao, Lei Ding, Kai Zhang, Guohua Tang, Xiao Han, and Yang Wei. 2024{\natexlab{b}}.
\newblock \href {https://aclanthology.org/2024.acl-long.220/} {{S}oft{D}edup: an efficient data reweighting method for speeding up language model pre-training}.
\newblock In \emph{Proceedings of the 62nd Annual Meeting of the Association for Computational Linguistics (Volume 1: Long Papers)}, pages 4011--4022, Bangkok, Thailand. Association for Computational Linguistics.

\bibitem[{Hendrycks et~al.(2021)Hendrycks, Burns, Basart, Zou, Mazeika, Song, and Steinhardt}]{mmlu}
Dan Hendrycks, Collin Burns, Steven Basart, Andy Zou, Mantas Mazeika, Dawn Song, and Jacob Steinhardt. 2021.
\newblock \href {https://openreview.net/forum?id=d7KBjmI3GmQ} {Measuring massive multitask language understanding}.
\newblock In \emph{International Conference on Learning Representations}.

\bibitem[{Kwiatkowski et~al.(2019)Kwiatkowski, Palomaki, Redfield, Collins, Parikh, Alberti, Epstein, Polosukhin, Devlin, Lee et~al.}]{nq}
Tom Kwiatkowski, Jennimaria Palomaki, Olivia Redfield, Michael Collins, Ankur Parikh, Chris Alberti, Danielle Epstein, Illia Polosukhin, Jacob Devlin, Kenton Lee, et~al. 2019.
\newblock Natural questions: a benchmark for question answering research.
\newblock \emph{Transactions of the Association for Computational Linguistics}, 7:453--466.

\bibitem[{Lai et~al.(2017)Lai, Xie, Liu, Yang, and Hovy}]{race}
Guokun Lai, Qizhe Xie, Hanxiao Liu, Yiming Yang, and Eduard Hovy. 2017.
\newblock \href {https://doi.org/10.18653/v1/D17-1082} {{RACE}: Large-scale {R}e{A}ding comprehension dataset from examinations}.
\newblock In \emph{Proceedings of the 2017 Conference on Empirical Methods in Natural Language Processing}, pages 785--794, Copenhagen, Denmark. Association for Computational Linguistics.

\bibitem[{Lin et~al.(2024)Lin, Gou, Gong, Liu, yelong shen, Xu, Lin, Yang, Jiao, Duan, and Chen}]{rho}
Zhenghao Lin, Zhibin Gou, Yeyun Gong, Xiao Liu, yelong shen, Ruochen Xu, Chen Lin, Yujiu Yang, Jian Jiao, Nan Duan, and Weizhu Chen. 2024.
\newblock \href {https://openreview.net/forum?id=0NMzBwqaAJ} {Not all tokens are what you need for pretraining}.
\newblock In \emph{The Thirty-eighth Annual Conference on Neural Information Processing Systems}.

\bibitem[{Liu et~al.(2025)Liu, Zheng, Muennighoff, Zeng, Dou, Pang, Jiang, and Lin}]{regmix}
Qian Liu, Xiaosen Zheng, Niklas Muennighoff, Guangtao Zeng, Longxu Dou, Tianyu Pang, Jing Jiang, and Min Lin. 2025.
\newblock \href {https://openreview.net/forum?id=5BjQOUXq7i} {Regmix: Data mixture as regression for language model pre-training}.
\newblock In \emph{The Thirteenth International Conference on Learning Representations}.

\bibitem[{Mihaylov et~al.(2018)Mihaylov, Clark, Khot, and Sabharwal}]{openbookqa}
Todor Mihaylov, Peter Clark, Tushar Khot, and Ashish Sabharwal. 2018.
\newblock \href {https://doi.org/10.18653/v1/D18-1260} {Can a suit of armor conduct electricity? a new dataset for open book question answering}.
\newblock In \emph{Proceedings of the 2018 Conference on Empirical Methods in Natural Language Processing}, pages 2381--2391, Brussels, Belgium. Association for Computational Linguistics.

\bibitem[{Muennighoff et~al.(2024)Muennighoff, Rush, Barak, Le~Scao, Tazi, Piktus, Pyysalo, Wolf, and Raffel}]{muennighoff2024scaling}
Niklas Muennighoff, Alexander Rush, Boaz Barak, Teven Le~Scao, Nouamane Tazi, Aleksandra Piktus, Sampo Pyysalo, Thomas Wolf, and Colin~A Raffel. 2024.
\newblock Scaling data-constrained language models.
\newblock \emph{Advances in Neural Information Processing Systems}, 36.

\bibitem[{Park et~al.(2023)Park, Georgiev, Ilyas, Leclerc, and Madry}]{ifscore}
Sung~Min Park, Kristian Georgiev, Andrew Ilyas, Guillaume Leclerc, and Aleksander Madry. 2023.
\newblock \href {https://proceedings.mlr.press/v202/park23c.html} {{TRAK}: Attributing model behavior at scale}.
\newblock In \emph{Proceedings of the 40th International Conference on Machine Learning}, volume 202 of \emph{Proceedings of Machine Learning Research}, pages 27074--27113. PMLR.

\bibitem[{Penedo et~al.(2024)Penedo, Kydl{\'\i}{\v{c}}ek, Lozhkov, Mitchell, Raffel, Von~Werra, Wolf et~al.}]{fineweb}
Guilherme Penedo, Hynek Kydl{\'\i}{\v{c}}ek, Anton Lozhkov, Margaret Mitchell, Colin Raffel, Leandro Von~Werra, Thomas Wolf, et~al. 2024.
\newblock The fineweb datasets: Decanting the web for the finest text data at scale.
\newblock \emph{arXiv preprint arXiv:2406.17557}.

\bibitem[{Penedo et~al.(2023)Penedo, Malartic, Hesslow, Cojocaru, Cappelli, Alobeidli, Pannier, Almazrouei, and Launay}]{refinedweb}
Guilherme Penedo, Quentin Malartic, Daniel Hesslow, Ruxandra Cojocaru, Alessandro Cappelli, Hamza Alobeidli, Baptiste Pannier, Ebtesam Almazrouei, and Julien Launay. 2023.
\newblock The refinedweb dataset for falcon llm: outperforming curated corpora with web data, and web data only.
\newblock \emph{arXiv preprint arXiv:2306.01116}.

\bibitem[{Peng et~al.(2025)Peng, Yang, Zeng, Lin, Liu, and Zhao}]{dataman}
Ru~Peng, Kexin Yang, Yawen Zeng, Junyang Lin, Dayiheng Liu, and Junbo Zhao. 2025.
\newblock \href {https://openreview.net/forum?id=eNbA8Fqir4} {Dataman: Data manager for pre-training large language models}.
\newblock In \emph{The Thirteenth International Conference on Learning Representations}.

\bibitem[{Qiu et~al.(2024)Qiu, Lv, Jin, Wang, Ning, Yu, Zhang, Chu, Qu, Peng et~al.}]{qiu2024wanjuan}
Jiantao Qiu, Haijun Lv, Zhenjiang Jin, Rui Wang, Wenchang Ning, Jia Yu, ChaoBin Zhang, Pei Chu, Yuan Qu, Runyu Peng, et~al. 2024.
\newblock Wanjuan-cc: A safe and high-quality open-sourced english webtext dataset.
\newblock \emph{arXiv preprint arXiv:2402.19282}.

\bibitem[{Rae et~al.(2021)Rae, Borgeaud, Cai, Millican, Hoffmann, Song, Aslanides, Henderson, Ring, Young et~al.}]{gopher}
Jack~W Rae, Sebastian Borgeaud, Trevor Cai, Katie Millican, Jordan Hoffmann, Francis Song, John Aslanides, Sarah Henderson, Roman Ring, Susannah Young, et~al. 2021.
\newblock Scaling language models: Methods, analysis \& insights from training gopher.
\newblock \emph{arXiv preprint arXiv:2112.11446}.

\bibitem[{Raffel et~al.(2020)Raffel, Shazeer, Roberts, Lee, Narang, Matena, Zhou, Li, and Liu}]{c4}
Colin Raffel, Noam Shazeer, Adam Roberts, Katherine Lee, Sharan Narang, Michael Matena, Yanqi Zhou, Wei Li, and Peter~J. Liu. 2020.
\newblock \href {http://jmlr.org/papers/v21/20-074.html} {Exploring the limits of transfer learning with a unified text-to-text transformer}.
\newblock \emph{Journal of Machine Learning Research}, 21(140):1--67.

\bibitem[{Sakaguchi et~al.(2020)Sakaguchi, Le~Bras, Bhagavatula, and Choi}]{winogrande}
Keisuke Sakaguchi, Ronan Le~Bras, Chandra Bhagavatula, and Yejin Choi. 2020.
\newblock Winogrande: An adversarial winograd schema challenge at scale.
\newblock In \emph{Proceedings of the AAAI Conference on Artificial Intelligence}, volume~34, pages 8732--8740.

\bibitem[{Sap et~al.(2019)Sap, Rashkin, Chen, Le~Bras, and Choi}]{siqa}
Maarten Sap, Hannah Rashkin, Derek Chen, Ronan Le~Bras, and Yejin Choi. 2019.
\newblock \href {https://doi.org/10.18653/v1/D19-1454} {Social {IQ}a: Commonsense reasoning about social interactions}.
\newblock In \emph{Proceedings of the 2019 Conference on Empirical Methods in Natural Language Processing and the 9th International Joint Conference on Natural Language Processing (EMNLP-IJCNLP)}, pages 4463--4473, Hong Kong, China. Association for Computational Linguistics.

\bibitem[{Soboleva et~al.(2023)Soboleva, Al-Khateeb, Myers, Steeves, Hestness, and Dey}]{slimpajama}
Daria Soboleva, Faisal Al-Khateeb, Robert Myers, Jacob~R Steeves, Joel Hestness, and Nolan Dey. 2023.
\newblock \href {https://huggingface.co/datasets/cerebras/SlimPajama-627B} {{SlimPajama: A 627B token cleaned and deduplicated version of RedPajama}}.

\bibitem[{Soldaini et~al.(2024)Soldaini, Kinney, Bhagia, Schwenk, Atkinson, Authur, Bogin, Chandu, Dumas, Elazar et~al.}]{dolma}
Luca Soldaini, Rodney Kinney, Akshita Bhagia, Dustin Schwenk, David Atkinson, Russell Authur, Ben Bogin, Khyathi Chandu, Jennifer Dumas, Yanai Elazar, et~al. 2024.
\newblock Dolma: An open corpus of three trillion tokens for language model pretraining research.
\newblock \emph{arXiv preprint arXiv:2402.00159}.

\bibitem[{Su et~al.(2024)Su, Ahmed, Lu, Pan, Bo, and Liu}]{su2024roformer}
Jianlin Su, Murtadha Ahmed, Yu~Lu, Shengfeng Pan, Wen Bo, and Yunfeng Liu. 2024.
\newblock Roformer: Enhanced transformer with rotary position embedding.
\newblock \emph{Neurocomputing}, 568:127063.

\bibitem[{Tirumala et~al.(2023)Tirumala, Simig, Aghajanyan, and Morcos}]{d4}
Kushal Tirumala, Daniel Simig, Armen Aghajanyan, and Ari Morcos. 2023.
\newblock \href {https://proceedings.neurips.cc/paper_files/paper/2023/file/a8f8cbd7f7a5fb2c837e578c75e5b615-Paper-Datasets_and_Benchmarks.pdf} {D4: Improving llm pretraining via document de-duplication and diversification}.
\newblock In \emph{Advances in Neural Information Processing Systems}, volume~36, pages 53983--53995. Curran Associates, Inc.

\bibitem[{Touvron et~al.(2023)Touvron, Lavril, Izacard, Martinet, Lachaux, Lacroix, Rozi{\`e}re, Goyal, Hambro, Azhar et~al.}]{llama}
Hugo Touvron, Thibaut Lavril, Gautier Izacard, Xavier Martinet, Marie-Anne Lachaux, Timoth{\'e}e Lacroix, Baptiste Rozi{\`e}re, Naman Goyal, Eric Hambro, Faisal Azhar, et~al. 2023.
\newblock Llama: Open and efficient foundation language models.
\newblock \emph{arXiv preprint arXiv:2302.13971}.

\bibitem[{Warner et~al.(2024)Warner, Chaffin, Clavié, Weller, Hallström, Taghadouini, Gallagher, Biswas, Ladhak, Aarsen, Cooper, Adams, Howard, and Poli}]{modernbert}
Benjamin Warner, Antoine Chaffin, Benjamin Clavié, Orion Weller, Oskar Hallström, Said Taghadouini, Alexis Gallagher, Raja Biswas, Faisal Ladhak, Tom Aarsen, Nathan Cooper, Griffin Adams, Jeremy Howard, and Iacopo Poli. 2024.
\newblock \href {http://arxiv.org/abs/2412.13663} {Smarter, better, faster, longer: A modern bidirectional encoder for fast, memory efficient, and long context finetuning and inference}.

\bibitem[{Weber et~al.(2024)Weber, Fu, Anthony, Oren, Adams, Alexandrov, Lyu, Nguyen, Yao, Adams, Athiwaratkun, Chalamala, Chen, Ryabinin, Dao, Liang, Ré, Rish, and Zhang}]{weber2024redpajama}
Maurice Weber, Daniel~Y. Fu, Quentin Anthony, Yonatan Oren, Shane Adams, Anton Alexandrov, Xiaozhong Lyu, Huu Nguyen, Xiaozhe Yao, Virginia Adams, Ben Athiwaratkun, Rahul Chalamala, Kezhen Chen, Max Ryabinin, Tri Dao, Percy Liang, Christopher Ré, Irina Rish, and Ce~Zhang. 2024.
\newblock Redpajama: an open dataset for training large language models.
\newblock \emph{NeurIPS Datasets and Benchmarks Track}.

\bibitem[{Welbl et~al.(2017)Welbl, Liu, and Gardner}]{sciq}
Johannes Welbl, Nelson~F. Liu, and Matt Gardner. 2017.
\newblock \href {https://doi.org/10.18653/v1/W17-4413} {Crowdsourcing multiple choice science questions}.
\newblock In \emph{Proceedings of the 3rd Workshop on Noisy User-generated Text}, pages 94--106, Copenhagen, Denmark. Association for Computational Linguistics.

\bibitem[{Wenzek et~al.(2020)Wenzek, Lachaux, Conneau, Chaudhary, Guzm{\'a}n, Joulin, and Grave}]{wenzek2020ccnet}
Guillaume Wenzek, Marie-Anne Lachaux, Alexis Conneau, Vishrav Chaudhary, Francisco Guzm{\'a}n, Armand Joulin, and {\'E}douard Grave. 2020.
\newblock Ccnet: Extracting high quality monolingual datasets from web crawl data.
\newblock In \emph{Proceedings of the Twelfth Language Resources and Evaluation Conference}, pages 4003--4012.

\bibitem[{Wettig et~al.(2024)Wettig, Gupta, Malik, and Chen}]{qurating}
Alexander Wettig, Aatmik Gupta, Saumya Malik, and Danqi Chen. 2024.
\newblock \href {https://openreview.net/forum?id=GLGYYqPwjy} {Qurating: Selecting high-quality data for training language models}.
\newblock In \emph{Forty-first International Conference on Machine Learning}.

\bibitem[{Xie et~al.(2023{\natexlab{a}})Xie, Pham, Dong, Du, Liu, Lu, Liang, Le, Ma, and Yu}]{doremi}
Sang~Michael Xie, Hieu Pham, Xuanyi Dong, Nan Du, Hanxiao Liu, Yifeng Lu, Percy~S Liang, Quoc~V Le, Tengyu Ma, and Adams~Wei Yu. 2023{\natexlab{a}}.
\newblock \href {https://proceedings.neurips.cc/paper_files/paper/2023/file/dcba6be91359358c2355cd920da3fcbd-Paper-Conference.pdf} {Doremi: Optimizing data mixtures speeds up language model pretraining}.
\newblock In \emph{Advances in Neural Information Processing Systems}, volume~36, pages 69798--69818. Curran Associates, Inc.

\bibitem[{Xie et~al.(2023{\natexlab{b}})Xie, Santurkar, Ma, and Liang}]{dsir}
Sang~Michael Xie, Shibani Santurkar, Tengyu Ma, and Percy~S Liang. 2023{\natexlab{b}}.
\newblock \href {https://proceedings.neurips.cc/paper_files/paper/2023/file/6b9aa8f418bde2840d5f4ab7a02f663b-Paper-Conference.pdf} {Data selection for language models via importance resampling}.
\newblock In \emph{Advances in Neural Information Processing Systems}, volume~36, pages 34201--34227. Curran Associates, Inc.

\bibitem[{Ye et~al.(2024)Ye, Liu, Sun, Zhou, Zhan, and Qiu}]{dmlaw}
Jiasheng Ye, Peiju Liu, Tianxiang Sun, Yunhua Zhou, Jun Zhan, and Xipeng Qiu. 2024.
\newblock Data mixing laws: Optimizing data mixtures by predicting language modeling performance.
\newblock \emph{arXiv preprint arXiv:2403.16952}.

\bibitem[{Yu et~al.(2024)Yu, Das, and Xiong}]{mates}
Zichun Yu, Spandan Das, and Chenyan Xiong. 2024.
\newblock \href {https://openreview.net/forum?id=6gzPSMUAz2} {{MATES}: Model-aware data selection for efficient pretraining with data influence models}.
\newblock In \emph{The Thirty-eighth Annual Conference on Neural Information Processing Systems}.

\bibitem[{Zellers et~al.(2019)Zellers, Holtzman, Bisk, Farhadi, and Choi}]{hellaswag}
Rowan Zellers, Ari Holtzman, Yonatan Bisk, Ali Farhadi, and Yejin Choi. 2019.
\newblock \href {https://doi.org/10.18653/v1/P19-1472} {{H}ella{S}wag: Can a machine really finish your sentence?}
\newblock In \emph{Proceedings of the 57th Annual Meeting of the Association for Computational Linguistics}, pages 4791--4800, Florence, Italy. Association for Computational Linguistics.

\bibitem[{Zhang et~al.(2025)Zhang, Zhong, Zhang, Chai, Wang, Zhuang, Bai, Jiantao, Cao, Fan, Yuan, Wang, and He}]{quad}
Chi Zhang, Huaping Zhong, Kuan Zhang, Chengliang Chai, Rui Wang, Xinlin Zhuang, Tianyi Bai, Qiu Jiantao, Lei Cao, Ju~Fan, Ye~Yuan, Guoren Wang, and Conghui He. 2025.
\newblock \href {https://openreview.net/forum?id=bMC1t7eLRc} {Harnessing diversity for important data selection in pretraining large language models}.
\newblock In \emph{The Thirteenth International Conference on Learning Representations}.

\bibitem[{Zhang et~al.(2024)Zhang, Luo, Yuan, and Yao}]{automathtext}
Yifan Zhang, Yifan Luo, Yang Yuan, and Andrew~C Yao. 2024.
\newblock Autonomous data selection with language models for mathematical texts.
\newblock In \emph{ICLR 2024 Workshop on Navigating and Addressing Data Problems for Foundation Models}.

\end{thebibliography}

\appendix
\newpage

\section{Ratings}
\label{app:redpajama}
\begin{figure*}[!tb]
    \centering
    \includegraphics[width=1\linewidth]{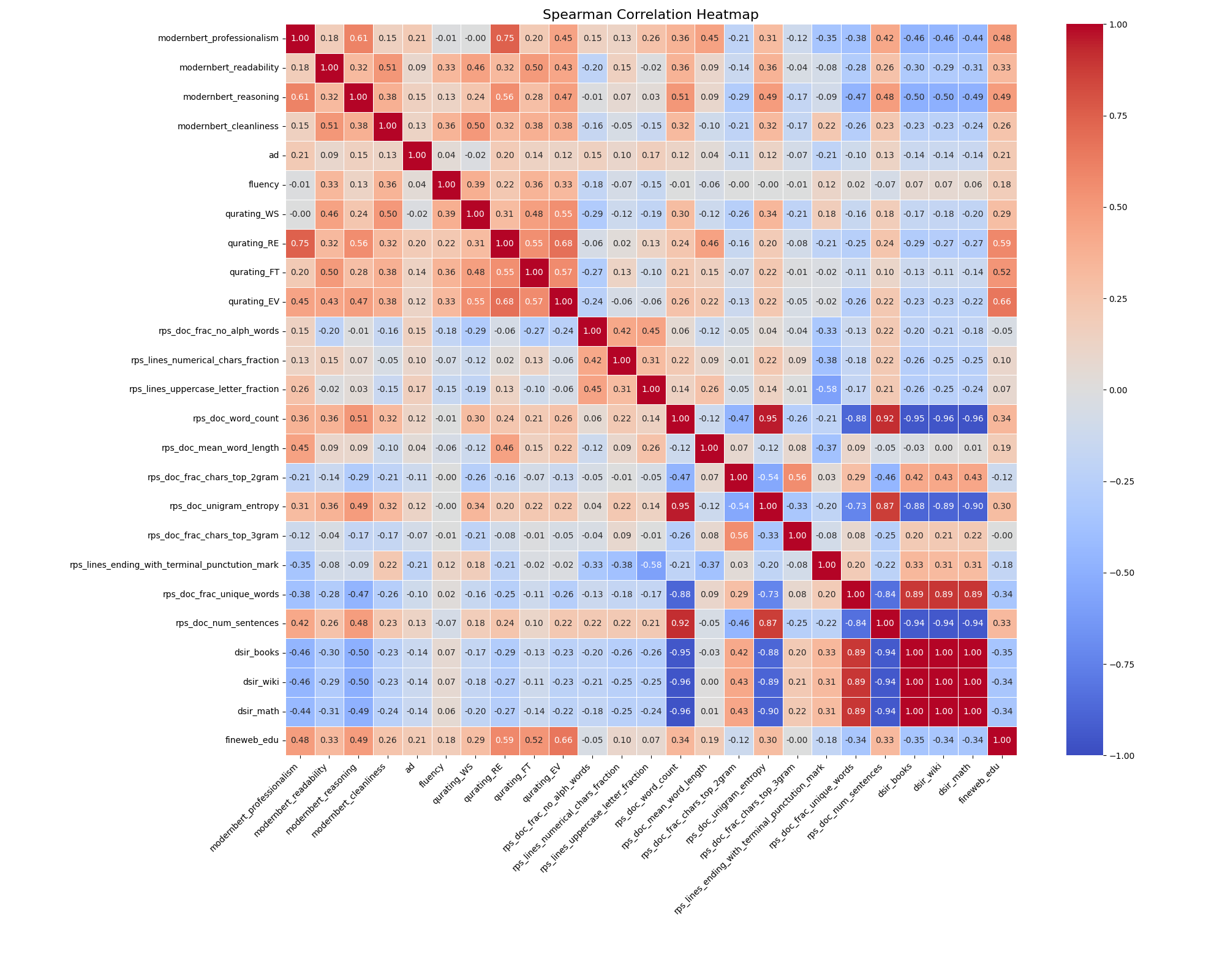}
    \caption{The spearman correlation heatmap among 25 quality metrics.}
    \label{fig:correlation_heatmap}
\end{figure*}

The full list of 25 raters utilized in this study, including 11 rule-based natural language quality signals, 3 data importance scores, and 11 model-based quality scores is shown in Table \ref{tab:full_list}.
The spearman correlation heatmap among 25 quality metrics is in Figure \ref{fig:correlation_heatmap}.

\begin{table*}
\centering
\large
\resizebox{\linewidth}{!}{
\begin{tabular}{@{} p{5cm} p{3cm} p{3cm} p{12cm} @{}}
\toprule
\textbf{Rater} &
  \textbf{Source} &
  \textbf{Type} &
  \textbf{Description} \\ \midrule
doc\_frac\_no\_alph\_words &
  \multirow{20}{*}{RedPajama}  &
  \multirow{20}{*}{\begin{tabular}[c]{@{}l@{}}Natural language \\ quality signals\end{tabular}} &
  The fraction of words that contain no alphabetical character. \\
doc\_mean\_word\_length &
   &
   &
  The mean length of words in the content after normalisation. \\
doc\_frac\_unique\_words &
   &
   &
  The fraction of unique words in the content. This is also known as the degeneracy of a text sample. \\
doc\_unigram\_entropy &
   &
   &
  The entropy of the unigram distribution of the content. This measures the diversity of the content and is computed using sum(-x / total * log(x / total)) where the sum is taken over counts of unique words in the normalised content. \\
doc\_word\_count &
   &
   &
  The number of words in the content after normalisation. \\
lines\_ending\_with \\ \_terminal\_punctution\_mark &
   &
   &
  Indicates whether a line ends with a terminal punctuation mark. A terminal punctation mark is defined as one of: ".", "!", "?", """. \\
lines\_numerical\_chars\_fraction &
   &
   &
  The ratio between the number of numerical characters and total number of characters in each line. This is based on the normalised content. \\
lines\_uppercase\_letter\_fraction &
   &
   &
  The ratio between the number of uppercase letters and total number of characters in each line. This is based on the raw text. \\
doc\_num\_sentences &
   &
   &
  The number of sentences in the content. This is calculated using the regular expression r'\textbackslash{}b{[}\textasciicircum{}.!?{]}+{[}.!?{]}*'. \\
doc\_frac\_chars\_top\_2gram &
   &
   &
  The fraction of characters in the top word 2-gram. \\
doc\_frac\_chars\_top\_3gram &
   &
   &
  The fraction of characters in the top word 3-gram. \\ \midrule
books\_importance &
  \multirow{9}{*}{DSIR} &
  \multirow{9}{*}{\begin{tabular}[c]{@{}l@{}}Data importance \\ scores\end{tabular}} &
  Given a bag of \{1,2\}-wordgram model trained on Books p, and a model trained on the source domain q, this is the logarithm of the ratio p(doc)/q(doc). \\
wikipedia\_importance &
   &
   &
  Given a bag of \{1,2\}-wordgram model trained on Wikipedia articles p, and a model trained on the source domain q, this is the logarithm of the ratio p(doc)/q(doc). \\
math\_importance &
   &
   &
  Given a bag of \{1,2\}-wordgram model trained on Math p, and a model trained on the source domain q, this is the logarithm of the ratio p(doc)/q(doc). \\ \midrule
Fineweb-edu &
  \multirow{2}{*}{Fineweb} &
  \multirow{22}{*}{\begin{tabular}[c]{@{}l@{}}Model-based \\ quality scores\end{tabular}} &
  This is a 110M BERT model for predicting educational value of a given text. \\
Advertisement &
  \multirow{4}{*}{WanjuanCC} &
   &
  This is a 110M BERT model for predicting whether a given text contains advertisement. \\
Fluency &
   &
   &
  This is a 110M BERT model for predicting whether a given text is fluent enough. \\
Required Expertise &
  \multirow{8}{*}{QuRating} &
   &
  This is a 1.3B Llama-style model for predicting whether a given text contains enough required expertise for understanding. \\
Writing Style &
   &
   &
  This is a 1.3B Llama-style model for predicting whether a given text has good writing style. \\
Facts and Trivia &
   &
   &
  This is a 1.3B Llama-style model for predicting whether a given text contains enough facts and trivia. \\
Educational Value &
   &
   &
  This is a 1.3B Llama-style model for predicting whether a given text contains enough required expertise for understanding. \\
Professionalism &
  \multirow{8}{*}{Ours} &
   &
  This is a 149M ModernBERT model for predicting Professionalism of a given text. \\
Readability &
   &
   &
  This is a 149M ModernBERT model for predicting Readability of a given text. \\
Reasoning &
   &
   
   &
  This is a 149M ModernBERT model for predicting Reasoning of a given text. \\
Cleanliness &
   &
   &
  This is a 149M ModernBERT model for predicting Cleanliness of a given text. \\ \bottomrule
\end{tabular}
}
\caption{A full list of all 25 raters used in this study.}
\label{tab:full_list}
\end{table*}

\begin{figure*}[!tb]
    \centering
    \includegraphics[width=0.8\linewidth]{figs/metarater_pca2_landscape.png}
    \caption{The loss landscape of proxy model losses over the first two principal components derived from the PCA of 25 quality scores.}
    \label{fig:loss_landscape}
\end{figure*}

\begin{table}
\centering
\begin{tabular}{@{}lc@{}}
\toprule
\textbf{Domain} & \multicolumn{1}{l}{\textbf{Weight}} \\ \midrule
CommonCrawl     & 52.20                               \\
C4              & 26.70                               \\
GitHub          & 5.20                                \\
Books           & 4.20                                \\
ArXiv           & 4.60                                \\
Wikipedia       & 3.80                                \\
StackExchange   & 3.30                                \\ \bottomrule
\end{tabular}
\caption{Exact domain weights (\%) of SlimPajama.}
\label{tab:weights}
\end{table}

\begin{table*}
\centering
\begin{tabular}{@{}lcc@{}}
\toprule
\multicolumn{1}{c}{\textbf{Characteristics}} & \textbf{Length}    & \textbf{Estimated Token Length} \\ \midrule
Min                                 & 5         & 6                      \\
Max                                 & 1,302,898 & 1,693,767              \\
Mean                                & 1029.8    & 1338.7                 \\
Median                              & 469.0     & 609.7                  \\
25\% Percentile                     & 204.0     & 265.2                  \\
50\% Percentile                     & 469.0     & 609.7                  \\
75\% Percentile                     & 957.0     & 1244.1                 \\
90\% Percentile                     & 1869.0    & 2429.7                 \\
95\% Percentile                     & 3033.0    & 3942.9                 \\
99\% Percentile                     & 9637.8    & 12529.1                \\ \bottomrule
\end{tabular}
\caption{Characteristics of SlimPajama. Length denotes the number of characters and Estimated Token Length denotes the estimated number of ModernBERT tokens.}
\label{tab:characteristics_slimpajama}
\end{table*}

\begin{table*}
\centering
\begin{tabular}{@{}ccccc@{}}
\toprule
\textbf{Model Type}                         & \textbf{Max context window} & \textbf{Accuracy}       & \textbf{F1}             & \textbf{Query Per Second} \\ \midrule
\multirow{3}{*}{Base (149M)}  & 8k                 & 92.82          & 89.73          & 325.82           \\
                                  & 4k                 & 92.66          & \textbf{90.12} & 420.53           \\
                                  & 2k                 & \textbf{92.93} & 89.89          & \textbf{478.98}  \\ \midrule
\multirow{3}{*}{Large (395M)} & 8k                 & 93.47          & 90.96          & 163.91           \\
                                  & 4k                 & 93.36          & 90.93          & 195.07           \\
                                  & 2k                 & \textbf{93.59} & \textbf{91.12} & \textbf{196.40}  \\ \bottomrule
\end{tabular}
\caption{Test results and average inference speed of ModernBERT models on one NVIDIA A800 GPU.}
\label{tab:train_inference_results}
\end{table*}

\begin{table*}
\centering
\begin{tabular}{@{}lccc@{}}
\toprule
\multicolumn{1}{c}{\textbf{Model}} & \textbf{Max context window} & \textbf{Accuracy} & \textbf{F1}    \\ \midrule
Base-Professionalism   & 4k                 & 93.78    & 91.57 \\
Base-Readability       & 4k                 & 94.13    & 87.47 \\
Base-Reasoning         & 4k                 & 96.32    & 89.59 \\
Base-Cleanliness       & 4k                 & 92.25    & 87.88 \\ \bottomrule
\end{tabular}
\caption{Test performance of rating models on the test set.}
\label{tab:modernbert}
\end{table*}

\section{Weights}
\label{app:weights}

\subsection{Meta-rater Weights}
\label{app:meta_rater_weights}
\begin{table*}
\centering
\begin{tabular}{@{}lcc@{}}
\toprule
\textbf{Rater}                                           & \textbf{Weight (\%)} & \textbf{Rank} \\ \midrule
Educational Value                               & 5.64        & 1    \\
doc\_frac\_no\_alph\_words                      & 4.93        & 2    \\
Fineweb-edu                                     & 4.93        & 2    \\
lines\_uppercase\_letter\_fraction              & 4.88        & 4    \\
Facts and Trivia                                & 4.77        & 5    \\
doc\_frac\_chars\_top\_3gram                    & 4.73        & 6    \\
lines\_ending\_with\_terminal\_punctution\_mark & 4.73        & 6    \\
doc\_frac\_chars\_top\_2gram                    & 4.71        & 8    \\
wikipedia\_importance                           & 4.69        & 9    \\
lines\_numerical\_chars\_fraction               & 4.60        & 10   \\
doc\_num\_sentences                             & 4.58        & 11   \\
math\_importance                                & 4.48        & 12   \\
Reasoning                                       & 4.44        & 13   \\
doc\_frac\_unique\_words                        & 4.32        & 14   \\
doc\_word\_count                                & 4.23        & 15   \\
doc\_unigram\_entropy                           & 4.22        & 16   \\
books\_importance                               & 4.14        & 17   \\
Professionalism                                 & 4.05        & 18   \\
Fluency                                         & 4.02        & 19   \\
Readability                                     & 3.93        & 20   \\
Required Expertise                              & 3.73        & 21   \\
Advertisement                                   & 3.68        & 22   \\
Cleanliness                                     & 1.17        & 23   \\
doc\_mean\_word\_length                         & 0.65        & 24   \\
Writing Style                                   & 0.05        & 25   \\ \bottomrule
\end{tabular}
\caption{Meta-rater learned weights for all raters.}
\label{tab:meta_rater_weights}
\end{table*}

\subsection{Domain Weights}
\label{app:domain_weights}
We list the exact domain weights in Table \ref{tab:weights}.

\section{PRRC Models}
\label{app:prrc-models}

\subsection{Annotation Model}
\label{app:annotation_model}
We selected 7 powerful LLMs as candidates for annotation: 
\textbf{Qwen-2.5-72B-Instruct}\footnote{\url{https://huggingface.co/Qwen/Qwen2.5-72B-Instruct}}, 
\textbf{Qwen-2-72B-Instruct}\footnote{\url{https://huggingface.co/Qwen/Qwen2-72B-Instruct}}, 
\textbf{Llama-3.3-70B-Instruct}\footnote{\url{https://huggingface.co/meta-llama/Llama-3.3-70B-Instruct}},
\textbf{Llama-3.1-70B-Instruct}\footnote{\url{https://huggingface.co/meta-llama/Llama-3.1-70B-Instruct}}, 
\textbf{Llama-3-70B-Instruct}\footnote{\url{https://huggingface.co/meta-llama/Meta-Llama-3-70B-Instruct}}, 
\textbf{gpt-4o}\footnote{\url{https://openai.com/index/hello-gpt-4o/}}, 
and \textbf{gpt-3.5-turbo-0125}\footnote{\url{https://openai.com/index/chatgpt/}}.
To determine the most suitable model, we constructed a validation dataset comprising 1,000 instances, drawing from a wide range of sources including Wikipedia, Books, Reddit, StackExchange, ArXiv, and CommonCrawl.
These candidate LLMs, along with gpt-4\footnote{\url{https://openai.com/index/gpt-4/}}, were then used to score the validation dataset based on the prompts outlined in Appendix \ref{app:annotation}.
To assess the consistency of the candidate models relative to gpt-4, we computed the Kendall tau correlation score between their respective scores.
The results, evaluated across four dimensions, consistently indicated that \textbf{Llama-3.3-70B-Instruct} outperformed the others, leading to its selection as our final annotation model.

\subsection{Prompts for Annotation}
\label{app:annotation}
The four prompts used to evaluate \textit{Professionalism}, \textit{Readability}, \textit{Reasoning}, and \textit{Cleanliness} are presented in Figures \ref{fig:prompt_pro}, \ref{fig:prompt_readability}, \ref{fig:prompt_reasoning}, and \ref{fig:prompt_cleanliness}. Using these prompts, \textbf{Llama-3.3-70B-Instruct} was tasked with annotating 1 million randomly sampled documents from SlimPajama, utilizing a maximum context window length of 128k tokens. 
After applying the necessary filtering and cleaning procedures, a total of 934,278 document-score pairs were retained. 
These pairs were subsequently divided into training, development, and test sets in an 8:1:1 ratio, resulting in 747,422 training pairs, 93,428 development pairs, and 93,428 test pairs.
Moreover, specific examples in annotated SlimPajama are provided:
\begin{itemize}
    \item Examples of documents rated from 0 to 5 in terms of \textit{Professionalism} are presented in Figures \ref{fig:pro_0}, \ref{fig:pro_1}, \ref{fig:pro_2}, \ref{fig:pro_3}, \ref{fig:pro_4}, and \ref{fig:pro_5}. These six documents include excerpts from a web page, a nursery rhyme, a magazine article, a popular science article, and two academic papers.
    \item Examples of documents rated from 0 to 5 in terms of \textit{Readability} are presented in Figures \ref{fig:read_0}, \ref{fig:read_1}, \ref{fig:read_2}, \ref{fig:read_3}, \ref{fig:read_4}, and \ref{fig:read_5}. These six documents consist of excerpts from one web page and five student essays.
    \item Examples of documents rated from 0 to 5 in terms of \textit{Reasoning} are presented in Figures \ref{fig:reas_0}, \ref{fig:reas_1}, \ref{fig:reas_2}, \ref{fig:reas_3}, \ref{fig:reas_4}, and \ref{fig:reas_5}. These six documents consist of excerpts from three web pages and three news.
    \item Examples of documents rated from 0 to 5 in terms of \textit{Cleanliness} are presented in Figures \ref{fig:clean_0}, \ref{fig:clean_1}, \ref{fig:clean_2}, \ref{fig:clean_3}, \ref{fig:clean_4}, and \ref{fig:clean_5}. These six documents consist of excerpts from six web pages.
\end{itemize}

\subsection{PRRC Models Training}
We selected ModernBERT \cite{modernbert} as the rating models for two key reasons. 
First, it demonstrates superior comprehension capabilities, supported by its ability to handle significantly longer context windows—up to 8,192 tokens, compared to the 512 tokens supported by BERT \cite{bert}. 
Second, it is more efficient for both training and inference due to its integration with FlashAttention-2 \cite{flashattention2}. 
To determine the most suitable version of ModernBERT for text evaluation, we conducted an analysis from two perspectives: data characteristics and model performance. 

\paragraph{Data}
We analyzed the key characteristics of the SlimPajama dataset, with the results summarized in Table \ref{tab:characteristics_slimpajama}. 
Notably, a context window of 512 tokens can only process less than half of the dataset, whereas a context window of 4,096 tokens is capable of handling over 95\% of the dataset.

\paragraph{Model}
We evaluated both ModernBERT-base and ModernBERT-large, testing them with maximum context window lengths of 8k, 4k, and 2k tokens. 
These models were fine-tuned for 10 epochs to assess their performance on the dimension of \textit{Professionalism} using a small subset of the training dataset (50,000 samples for training and 10,000 samples for test). 
Additionally, we measured the average inference speed on a single NVIDIA A800 GPU, using the largest possible batch sizes.
As shown in Table \ref{tab:train_inference_results}, among the base models, the 4k version achieved the highest F1 score, making it the optimal choice within the Base model category. 
Furthermore, its inference speed was 28\% faster than the 8k model and only 12\% slower than the 2k model.

\paragraph{Training}
Ultimately, we selected \textbf{ModernBERT-base-4k} to evaluate four dimensions of text quality. Each model was fine-tuned for 10 epochs, and the performance on test set is presented in Table \ref{tab:modernbert}.

\section{Pre-training}
\label{app:training}
The specific architectures of all pre-trained models in this work are shown in Table \ref{tab:architecture}.
In all models, we employ the LLaMA tokenizer \cite{llama} with a vocabulary size of 32,000. The MLP ratio is configured to 8/3, the RoPE base is set to 10,000, and the maximum context length is fixed at 1,024 tokens.
Each model was trained on 32x NVIDIA A800 GPU, employing a global batch size of 4,194,304 tokens.
The learning rate was set to $5 \times 10^{-5}$, and the Adam optimizer was employed with hyperparameters ($\beta_1=0.9, \beta_2=0.95, \epsilon=10^{-8}$).

\begin{table*}[!tb]
\centering
\resizebox{\linewidth}{!}{
\begin{tabular}{@{}lcccccc@{}}
\toprule
\multicolumn{1}{c}{\textbf{Hyperparameter}} & \multicolumn{1}{c}{\textbf{18M (proxy model)}} & \textbf{178M}        & \textbf{407M}        & \textbf{1.3B}          & \textbf{3.3B}          & \textbf{7.2B}            \\ \midrule
Hidden Dimension Size         & 256 & 896 & 1,280 & 2,048 & 2,560 & 4,096 \\
Number of Layers              & 2   & 12  & 16    & 24    & 40    & 32    \\
Number of Attention Heads     & 4   & 7   & 10    & 16    & 20    & 32    \\
Number of KV  Heads  & 4   & 7   & 10    & 16    & 20    & 8     \\
Number of Total Parameters         & 18,089,216              & 178,476,928 & 407,020,800 & 1,345,423,360 & 3,335,989,760 & 7,241,732,096 \\
Consumed Tokens (B) & 0.5 & 3   & 6     & 30    & 100   & 150   \\
Pre-training Time (h)         & 0.1 & 0.3 & 0.5   & 14.0  & 129.0 & 284.0 \\ \bottomrule
\end{tabular}
}
\caption{Architectures of pre-trained decoder-only model.}
\label{tab:architecture}
\end{table*}

\begin{table*}
\centering
\begin{tabular}{@{}lc@{}}
\toprule
\textbf{Task}           & \textbf{Number} \\ \midrule
ARC-E          & 15     \\ \midrule
ARC-C          & 15     \\ \midrule
SciQ           & 2      \\ \midrule
HellaSwag      & 6      \\ \midrule
SIQA           & 10     \\ \midrule
WinoGrande     & 15     \\ \midrule
RACE           & 2      \\ \midrule
OpenbookQA     & 10     \\ \midrule
\end{tabular}
\caption{Number of demonstrations in in-context learning used for each downstream task.}
\label{tab:icl}
\end{table*}

\section{Evaluation}
\label{app:icl}
The number of randomly selected demonstrations for few-shot in-context learning for each task is listed in Table \ref{tab:icl}.

\section{Cost Analysis}
\label{app:cost}
We use Equation \ref{eq:train} to approximate FLOPs for training on transformer-style models.
\begin{equation}
    \label{eq:train}
    F_\text{train} = 6 \times L \times \ H^2 \times T \times |D_\text{train}| \times E 
\end{equation}
where $L$ denotes the number of model layers, $H$ denotes the hidden size, $T$ denotes number of tokens per sample, $|D_\text{train}|$ denotes the number of training samples, and $E$ denotes the number of training epochs.
\par
Similarly, the inference FLOPs can be approximated as:
\begin{equation}
    \label{eq:infer}
    F_\text{infer} = 2  \times L \times \ H^2 \times T \times |D_\text{infer}|
\end{equation}
where $|D_\text{infer}|$ denotes the number of samples to infer on.

\section{Distribution of Raters}
\label{app:distribution}
\begin{figure*}
    \centering
    \includegraphics[width=1.0\linewidth]{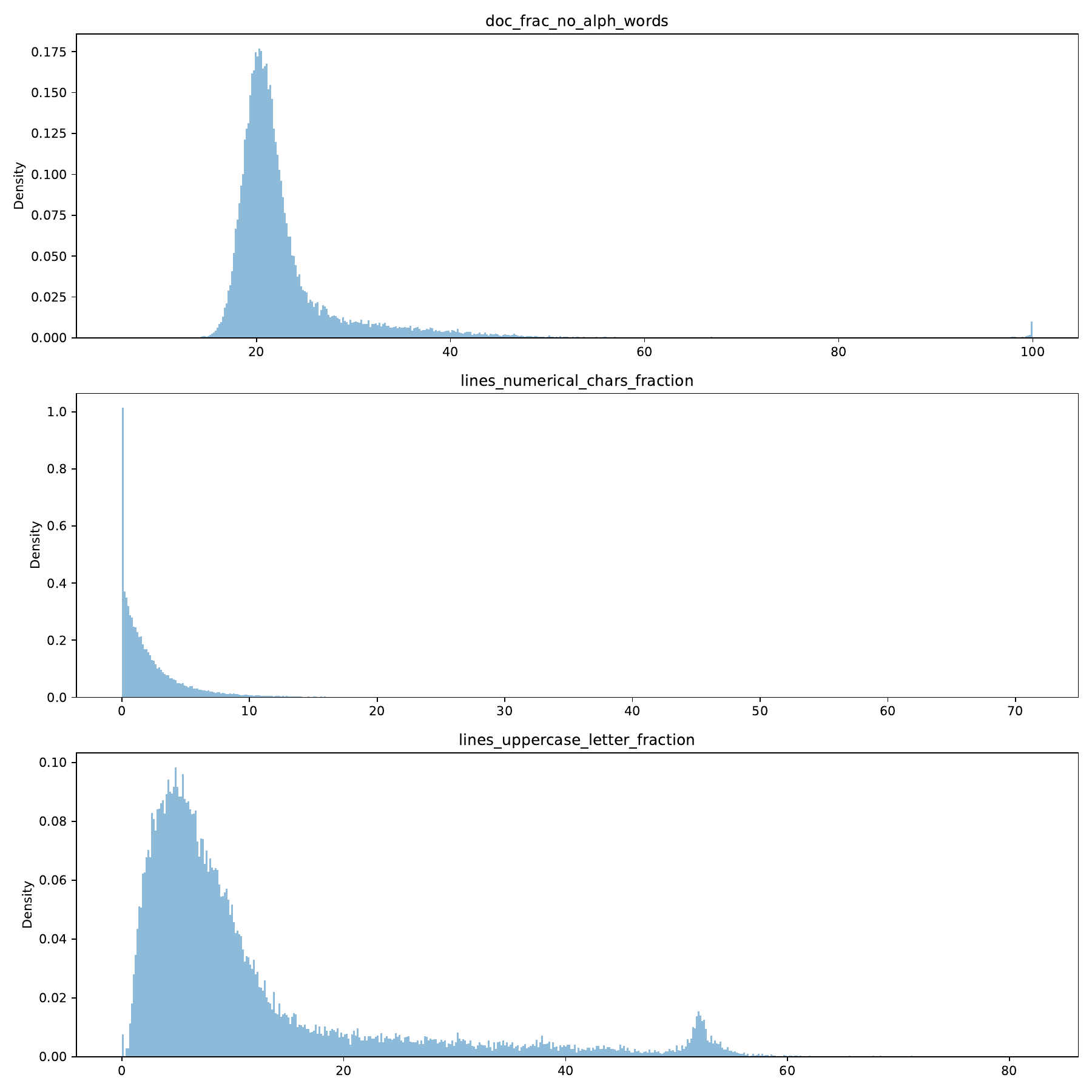}
    \caption{Distribution of natural language quality signals (Part 1/4).}
    \label{fig:dist_1}
\end{figure*}

\begin{figure*}
    \centering
    \includegraphics[width=1.0\linewidth]{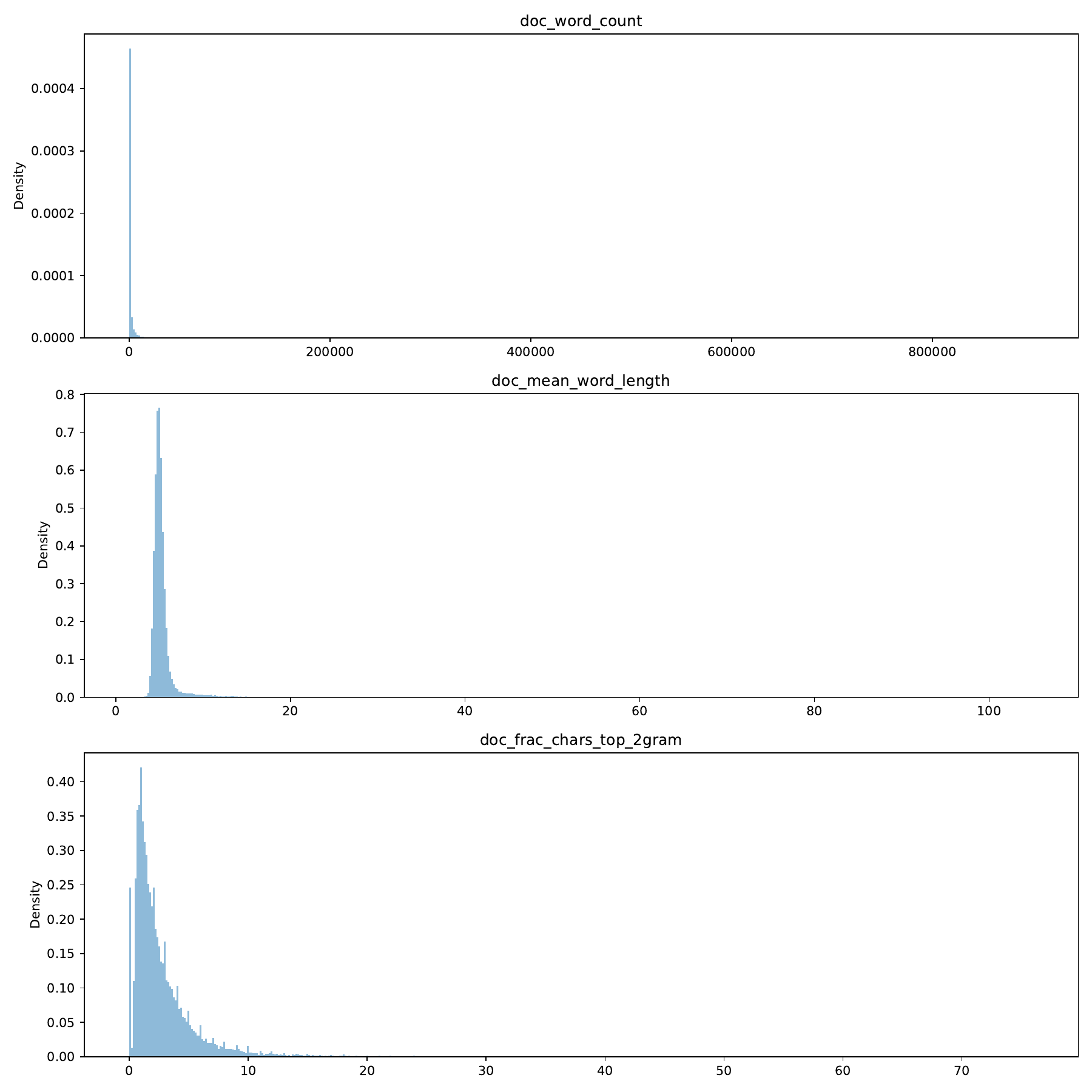}
    \caption{Distribution of natural language quality signals (Part 2/4).}
    \label{fig:dist_2}
\end{figure*}

\begin{figure*}
    \centering
    \includegraphics[width=1.0\linewidth]{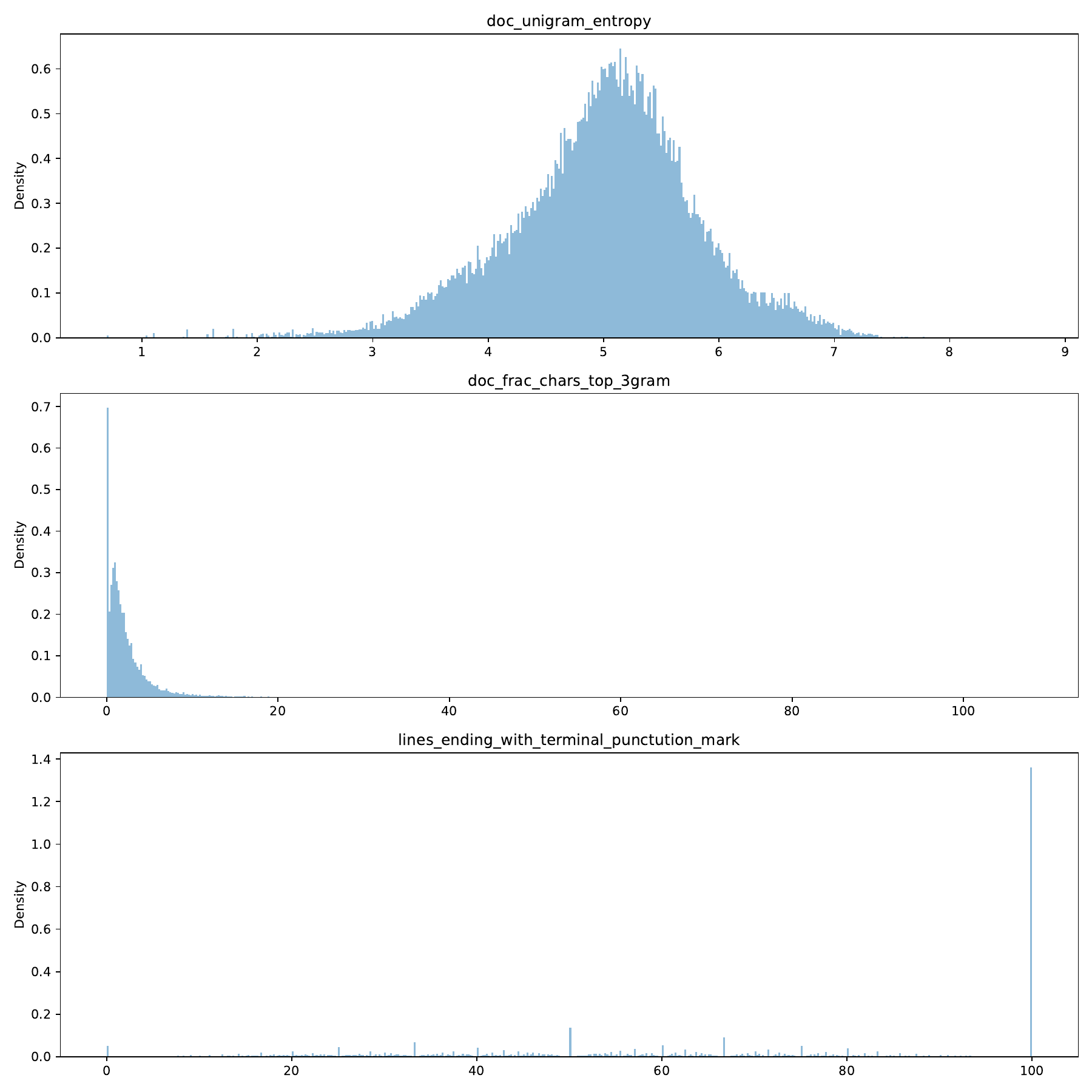}
    \caption{Distribution of natural language quality signals (Part 3/4).}
    \label{fig:dist_3}
\end{figure*}

\begin{figure*}
    \centering
    \includegraphics[width=1.0\linewidth]{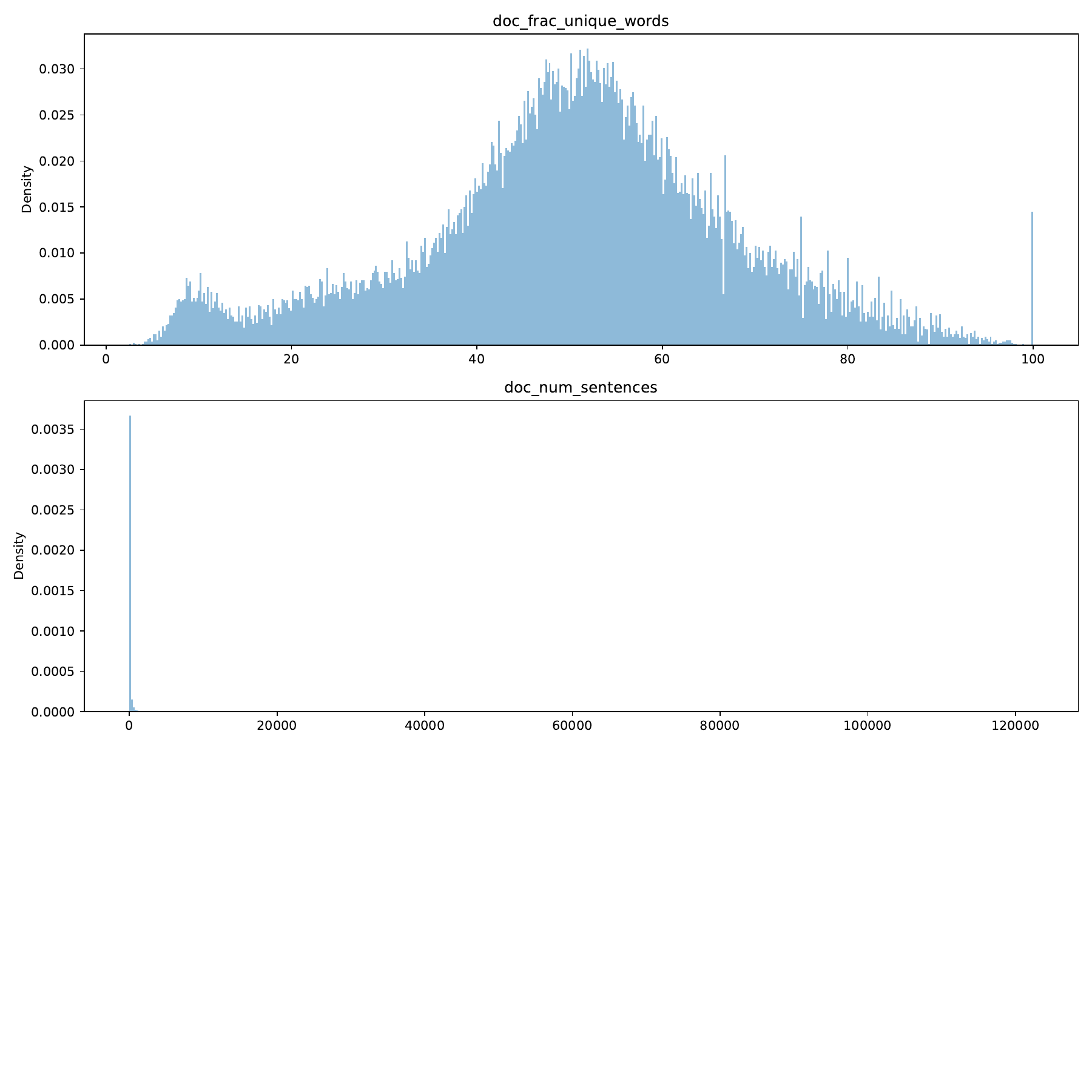}
    \caption{Distribution of natural language quality signals (Part 4/4).}
    \label{fig:dist_4}
\end{figure*}

\begin{figure*}
    \centering
    \includegraphics[width=1.0\linewidth]{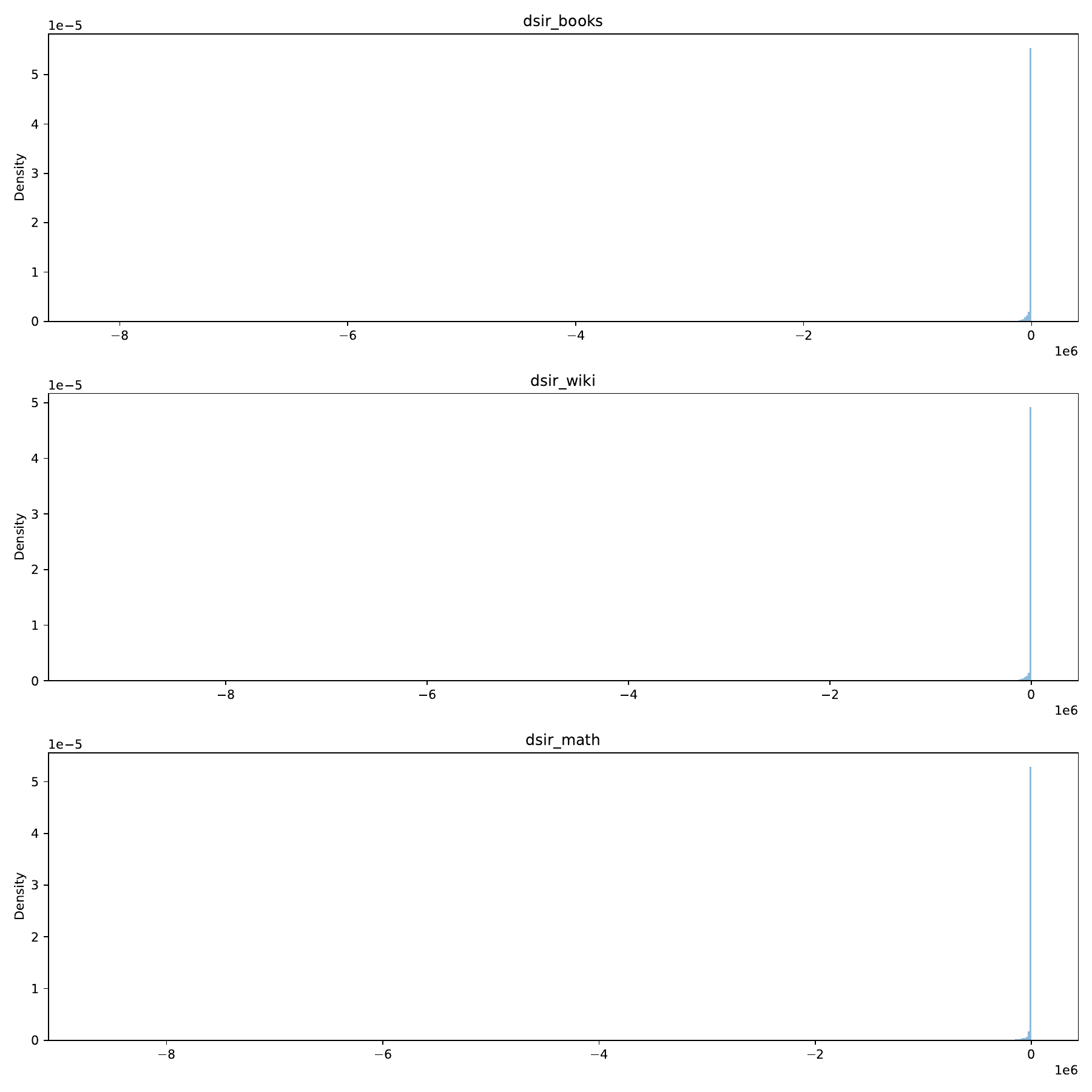}
    \caption{Distribution of data importance scores.}
    \label{fig:dist_5}
\end{figure*}

\begin{figure*}
    \centering
    \includegraphics[width=1.0\linewidth]{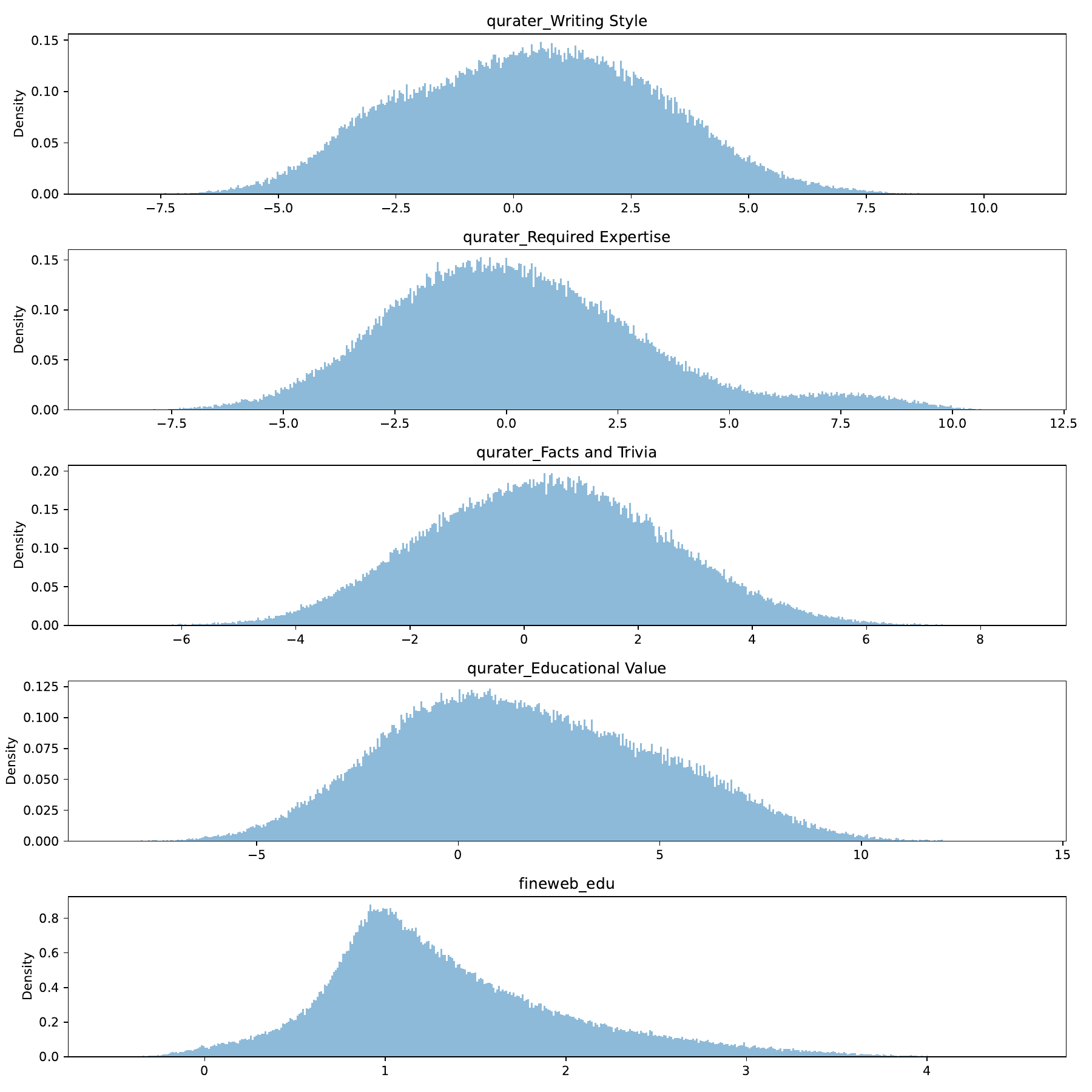}
    \caption{Distribution of model-based quality scores (Part 1/2).}
    \label{fig:dist_6}
\end{figure*}

\begin{figure*}
    \centering
    \includegraphics[width=1.0\linewidth]{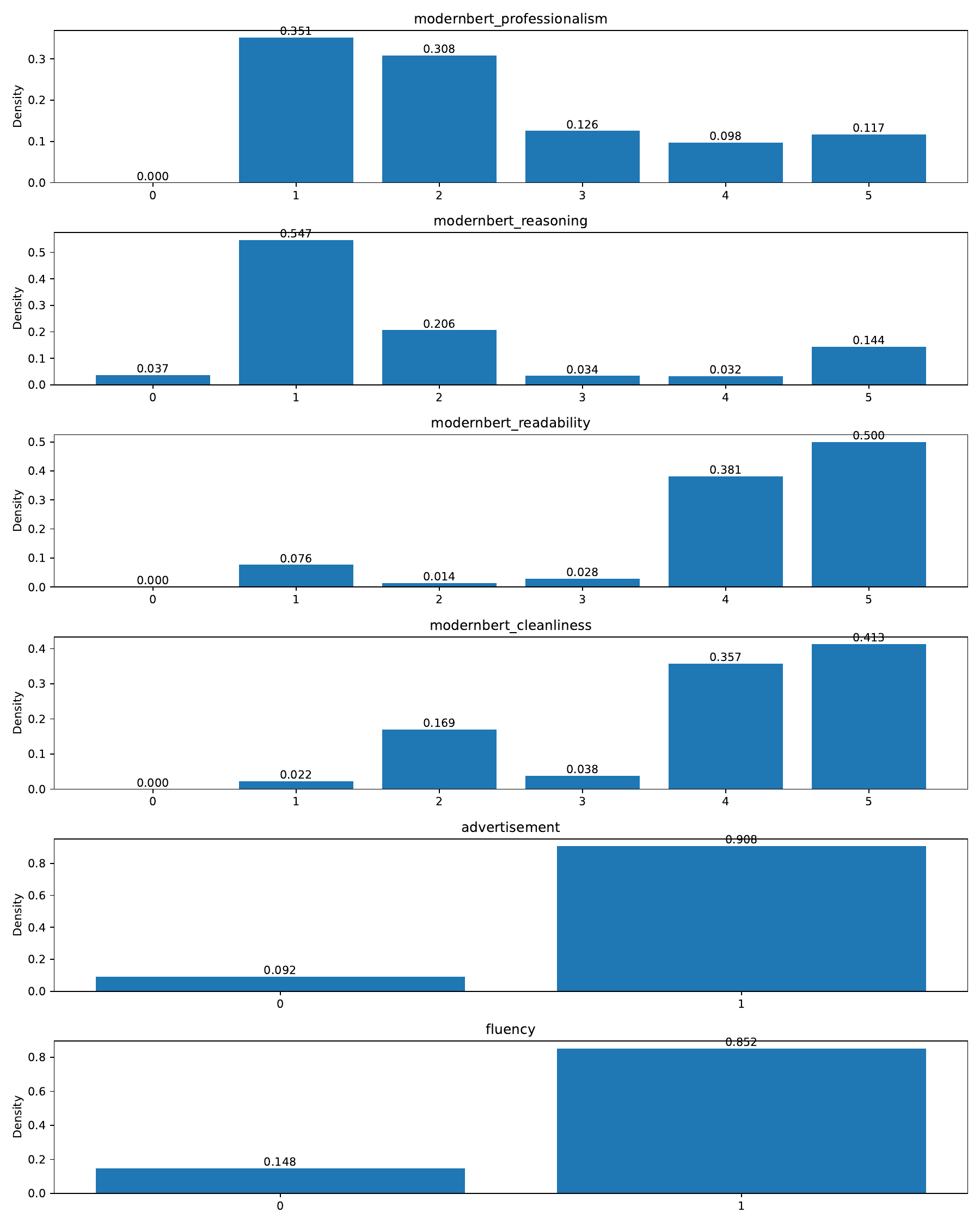}
    \caption{Distribution of model-based quality scores (Part 2/2).}
    \label{fig:dist_7}
\end{figure*}

The distribution of 11 rule-based natural language quality signals is shown in Figures \ref{fig:dist_1}, \ref{fig:dist_2}, \ref{fig:dist_3}, and \ref{fig:dist_4}.
The distribution of three data importance scores is shown in Figure \ref{fig:dist_5}.
The distribution of 11 model-based quality scores is shown in Figures \ref{fig:dist_6} and \ref{fig:dist_7}.

\begin{figure*}
    \centering
    \begin{tcolorbox}[promptboxstyle, title=Professionalism]
\textbf{CONTEXT}
I am a data scientist interested in exploring data in the pre-training stage of large language models.

\textbf{OBJECTIVE}
You are an expert evaluator. Below is an extract from a text source such as a web page, book, academic paper, Github, Wikipedia, or StackExchange. Evaluate the PROFESSIONALISM of the text, that is, the degree of expertise and prerequisite knowledge required to comprehend it, using the additive 5-point rating system described below. Your evaluation should be based on the depth, accuracy, and accessibility of the content, without considering the writing style, grammar, spelling, or punctuation in your rating.

Points are accumulated based on the satisfaction of each criterion:

- Add 1 point if the text is relatively simple and requires minimal technical knowledge or expertise to understand. The text might include nursery rhymes, children's books, or other basic content that is accessible to a broad audience. The information provided is straightforward and does not delve into complex concepts or specialized topics. 

- Add another point if the text is somewhat more complex and might require a basic level of specialized knowledge to comprehend fully. Examples could include popular books, popular science articles, or novels. The content delves a little deeper into the subject matter, but it remains accessible to a reasonabl0y broad audience.

- Award a third point if the text falls in the middle of the spectrum, requiring some degree of expertise to understand but not being overly complex or specialized. The content might encompass more advanced books, detailed articles, or introductions to complex topics. It provides a decent level of depth and detail, but it does not require an extensive background in the subject matter to understand.

- Grant a fourth point if the text is complicated and requires a significant level of expertise and technical knowledge. Examples might include academic papers, advanced textbooks, or detailed technical reports. The content is detailed and accurate, but it could be inaccessible to those without a strong background in the subject matter.

- Bestow a fifth point if the text is extremely high in professionalism, requiring a high degree of subject matter expertise and prerequisite knowledge. The text is likely limited to those with advanced understanding or experience in the field, such as advanced academic papers, complex technical manuals, or patents. The content is deep, accurate, and insightful, but largely inaccessible to those without a significant background in the topic.

Here are three aspects that should NOT influence your judgement: 
(1) The specific language the text is written in.

(2) The length of text.

(3) Usage of placeholders for data privacy or safety.

\textbf{STYLE}
A formal and clear text including score and reason.

\textbf{TONE}
professional, objective, formal, and clear.

\textbf{AUDIENCE}
Data scientists and other professionals interested in data for large language models.

\textbf{RESPONSE}
After examining the text, briefly justify your total score, up to 100 words. Conclude with the score using the format: "Professionalism:{total points}".

Here is the text:  \{TEXT\}
    \end{tcolorbox}
    \caption{Prompt for evaluating \textit{Professionalism} of texts.}
    \label{fig:prompt_pro}
\end{figure*}

\begin{figure*}
    \centering
    \begin{tcolorbox}[promptboxstyle, title=Readability]
\textbf{CONTEXT}
I am a data scientist interested in exploring data in the pre-training stage of large language models.

\textbf{OBJECTIVE}
You are an expert evaluator. Below is an extract from a text source such as a web page, book, academic paper, Github, Wikipedia, or StackExchange.  Evaluate whether the page has a high READABILITY using the additive 5-point rating system described below. 

Points are accumulated based on the satisfaction of each criterion:

- Add 1 point if the text is somewhat readable but contains significant issues with clarity or coherence. It might include complex vocabulary or sentence structures that require advanced reading skills, or it might have numerous grammar and spelling errors.

- Add another point if the text is generally clear and coherent, but there are sections that are difficult to comprehend due to occasional grammar, spelling errors, or convoluted sentence structures.

- Award a third point if the text is clear and coherent for the most part, using appropriate vocabulary and sentence structures that are easy to understand. Minor issues with grammar or spelling might still be present.

- Grant a fourth point if the text is very clear and coherent, with very few or no errors in grammar and spelling. The text uses proper punctuation, vocabulary, and sentence structures that are easy to follow and understand.

- Bestow a fifth point if the text is outstanding in its clarity and coherence. It uses language and sentence structures that are easy to comprehend, while also conveying ideas and nuances effectively. Minor errors in grammar, spelling, and punctuation are allowed, but they should not interfere with the overall understanding of the text.

Here are three aspects that should NOT influence your judgement: 

(1) The specific language the text is written in.

(2) The length of text.

(3) Usage of placeholders for data privacy or safety.

\textbf{STYLE}
A formal and clear text including score and reason.

\textbf{TONE}
professional, objective, formal, and clear.

\textbf{AUDIENCE}
Data scientists and other professionals interested in data for large language models.

\textbf{RESPONSE}
After examining the text, briefly justify your total score, up to 100 words. Conclude with the score using the format: "Readability:{total points}".

Here is the text:  \{TEXT\}
    \end{tcolorbox}
    \caption{Prompt for evaluating \textit{Readability} of texts.}
    \label{fig:prompt_readability}
\end{figure*}

\begin{figure*}
    \centering
    \begin{tcolorbox}[promptboxstyle, title=Reasoning]
\textbf{CONTEXT}
I am a data scientist interested in exploring data in the pre-training stage of large language models.

\textbf{OBJECTIVE}
You are an expert evaluator. Below is an extract from a text source such as a web page, book, academic paper, Github, Wikipedia, or StackExchange.  Evaluate whether the page has a high REASONING using the additive 5-point rating system described below.  

Points are accumulated based on the satisfaction of each criterion:

- Add 1 point if the content contains preliminary elements of reasoning, possibly involving a single causal relationship or simple logical judgments, but lacks in-depth analysis (e.g., presenting a viewpoint without supporting evidence or detailed explanations).

- Add another point if the content demonstrates basic reasoning ability, incorporating some logical relationships that require the reader to engage in a certain level of thought. This may involve simple argumentative structures or examples, but the analysis remains superficial (e.g., providing a problem and a straightforward solution with some examples but lacking depth).

- Award a third point if the content exhibits a good level of reasoning complexity, involving multiple reasoning steps that require more complex thought from the reader. The reader should be able to identify several interrelated arguments and may include some depth of analysis (e.g., analyzing how different factors influence an outcome or comparing various viewpoints).

- Grant a fourth point if the content has a high level of reasoning complexity, including multi-layered logical reasoning and in-depth analysis. The reader needs to engage in complex thinking and can identify multiple interconnected arguments while conducting a comprehensive evaluation (e.g., analyzing multiple variables or assessing the pros and cons of different solutions).

- Bestow a fifth point if the content excels in reasoning complexity, demanding deep analysis and innovative thinking from the reader. The reasoning process is complex and multi-dimensional, involving interdisciplinary knowledge, requiring the reader to integrate various pieces of information to make comprehensive judgments (e.g., discussing complex mathematical models, designing optimization algorithms, or engaging in high-level strategic thinking).

Here are three aspects that should NOT influence your judgement: 

(1) The specific language the text is written in.

(2) The length of text.

(3) Usage of placeholders for data privacy or safety.

\textbf{STYLE}
A formal and clear text including score and reason.

\textbf{TONE}
professional, objective, formal, and clear.

\textbf{AUDIENCE}
Data scientists and other professionals interested in data for large language models.

\textbf{RESPONSE}
After examining the text, briefly justify your total score, up to 100 words. Conclude with the score using the format: "Reasoning:{total points}".

Here is the text:  \{TEXT\}
    \end{tcolorbox}
    \caption{Prompt for evaluating \textit{Reasoning} of texts.}
    \label{fig:prompt_reasoning}
\end{figure*}

\begin{figure*}
    \centering
    \begin{tcolorbox}[promptboxstyle, title=Cleanliness]
\textbf{CONTEXT}
I am a data scientist interested in exploring data in the pre-training stage of large language models.

\textbf{OBJECTIVE}
You are an expert evaluator. Below is an extract from a text source such as a web page, book, academic paper, Github, Wikipedia, or StackExchange.  Evaluate whether the page has a high CLEANLINESS using the additive 5-point rating system described below.  

Points are accumulated based on the satisfaction of each criterion:

- A score of 1 indicates serious issues that affect fluency.

- A score of 2 indicates the text has obvious problems that affect fluency.

- A score of 3 means that the text has some problems but does not seriously affect reading fluency.

- A score of 4 indicates the text has minor problems but does not affect reading.

- A score of 5 means points means that the text is perfect on every criteria.

High cleanliness is defined by the following four criteria, please score each of the four criteria on a 5-point scale:

- Correct formatting: The text should appear to be edited by a human, rather than extracted by a machine, with no inappropriate characters.

- Appropriate content: The text should not contain links, advertisements, or other irrelevant text that affects reading. The effective content of the text is long enough to extract a clear structure and theme.

- Completeness Content: The body of the article consists of complete sentences written naturally by humans, rather than phrases and lists, containing opinions, facts or stories.

However, if there is a \$TRUNCATED\$ symbol at the end, it should be considered as a manual article ending flag set by the author, and there is no need to consider completeness.

Here are three aspects that should NOT influence your judgement: 

(1) The specific language the text is written in.

(2) The length of text.

(3) Usage of placeholders for data privacy or safety.

\textbf{STYLE}
A formal and clear text including score and reason.

\textbf{TONE}
professional, objective, formal, and clear.

\textbf{AUDIENCE}
Data scientists and other professionals interested in data for large language models.

\textbf{RESPONSE}
After examining the text, briefly justify your total score, up to 100 words. Conclude with the score using the format:

Cleanliness: {Overall score}

Correct Formatting: {Correct Formatting score}

Appropriate Content: {Appropriate Content score}

Completeness Content: {Completeness Content score}

Here is the text:  \{TEXT\}
    \end{tcolorbox}
    \caption{Prompt for evaluating \textit{Cleanliness} of texts.}
    \label{fig:prompt_cleanliness}
\end{figure*}

\begin{figure*}
    \centering
    \begin{tcolorbox}[exampleboxstyle, title=Professionalism Score: 0]
I inset to apply my knowledge about fine art geometric sequences of visual sense. Artists, welcome to the tough world of science! I have seen at least 25 artists complaining that scientists never accept that their work could help science or that it was highly significant! I myself wrote a few highly critical articles on the work of artists (1,2,3,4). Why do I do this? Am I against art and these artists? Do scientists think that art is inferior to science? Definitely not! I respect art as much as I respect science.
    \end{tcolorbox}
    \caption{An example text excerpt with \textit{Professionalism} score of 0.}
    \label{fig:pro_0}
\end{figure*}

\begin{figure*}
    \centering
    \begin{tcolorbox}[exampleboxstyle, title=Professionalism Score: 1]
The settings of the story is very desert like what I mean by that is its the desert and its hot and he really had to push his self to make it. The condition at the rode are very rough and has a lot of shape turn it in. Apart in the story he gets a boost of energy and pedal as hard as he can down a hill and then relaxes. This is the setting and is summary of the story…
    \end{tcolorbox}
    \caption{An example text excerpt with \textit{Professionalism} score of 1.}
    \label{fig:pro_1}
\end{figure*}

\begin{figure*}
    \centering
    \begin{tcolorbox}[exampleboxstyle, title=Professionalism Score: 2]
This article is based on an expert interview with Kent Bry, conducted by wikiHow Staff Editors. Kent Bry is a certified ski and
snowboarding instructor and the director of Adventure Ski \& Snowboard, a school based in the San Diego, California metro area. With over
50 years of skiing and snowboarding performance and instruction experience, Kent is certified by the Professional Ski Instructors of
America(PSIA). Adventure Ski \& Snowboard is a member of the PSIA and the American Association of Snowboard Instructors (AASI).
Kent holds a BS in Recreational Therapy from San Diego State University and is also a California-registered recreational therapist. This
article has been viewed 1,381 times. You've never been skiing before and you have an upcoming trip to the slopes...
    \end{tcolorbox}
    \caption{An example text excerpt with \textit{Professionalism} score of 2.}
    \label{fig:pro_2}
\end{figure*}

\begin{figure*}
    \centering
    \begin{tcolorbox}[exampleboxstyle, title=Professionalism Score: 3]
What Is Dopamine? Dopamine is a type of neurotransmitter. Your body makes it, and your nervous system uses it to send messages
between nerve cells. That's why it's sometimes called a chemical messenger. Dopamine plays a role in how we feel pleasure. It's a big part of our unique human ability to think and plan. It helps us strive, focus, and find things interesting. Your body spreads it along four major pathways in the brain. Like most other systems in the body, you don't notice it (or maybe even know about it) until there's a problem. Too much or too little of it can lead to a vast range of health issues. Some are serious, like Parkinson's disease. Others are much less dire. Dopamine Basics It's made in the brain through a two-step process. First, it changes the amino acid tyrosine to a substance called dopa, and then into dopamine. It affects many parts of your behavior and physical functions, such as: Role in Mental Health It's hard to pinpoint a single cause of most mental health disorders and challenges. But they're often linked to too much or too little dopamine in different parts of the brain. Examples include: Schizophrenia. Decades ago, researchers believed that symptoms stemmed from a hyperactive dopamine system...
    \end{tcolorbox}
    \caption{An example text excerpt with \textit{Professionalism} score of 3.}
    \label{fig:pro_3}
\end{figure*}

\begin{figure*}
    \centering
    \begin{tcolorbox}[exampleboxstyle, title=Professionalism Score: 4]
All transformers have the same primary components: Tokenizers, which convert text into tokens. A single embedding layer, which converts
tokens and positions of the tokens into vector representations. Transformer layers, which carry out repeated transformations on the vector representations, extracting more and more linguistic information. These consist of alternating attention and feedforward layers.
(optional) Un-embedding layer, which converts the final vector representations back to a probability distribution over the tokens.
Transformer layers can be one of two types, encoder and decoder. In the original paper both of them were used, while later models included only one type of them. BERT is an example of an encoder-only model; GPT are decoder-only models.
Input The input text is parsed into tokens by a tokenizer, most often a byte pair encoding tokenizer, and each token is converted into a vector via looking up from a word embedding table. Then, positional information of the token is added to the word embedding....
    \end{tcolorbox}
    \caption{An example text excerpt with \textit{Professionalism} score of 4.}
    \label{fig:pro_4}
\end{figure*}

\begin{figure*}
    \centering
    \begin{tcolorbox}[exampleboxstyle, title=Professionalism Score: 5]
by the fact that elements of the vRKHS $\mathrm{G}$ defined by the kernel $K(x, x') = k(x,x')$ $\mathrm{Id}_\mathcal{H}$ can be interpreted as Hilbert--Schmidt operators on $\mathcal{H}$. We again recall that the space of Hilbert--Schmidt operators $\mathcal{H}$ is isometrically
isomorphic to the tensor product space $\mathcal{H} \otimes \mathcal{H}$ via an identification of rank-one operators as elementary tensors.
We will use the latter to state the result, since a formulation in this way is more natural...
    \end{tcolorbox}
    \caption{An example text excerpt with \textit{Professionalism} score of 5.}
    \label{fig:pro_5}
\end{figure*}

\begin{figure*}
    \centering
    \begin{tcolorbox}[exampleboxstyle, title=Readability Score: 0]
"Friday, May 23, 2008 at 10:00 a.m. 2. Adoption of Minutes of the April 22, 2008 Urban Forestry Council Regular Meeting (Explanatory Document: Draft Minutes of the April 22, 2008 Regular Meeting) (Discussion and Action). 3. Informational Report from the Mayor's Office Director of City Greening on urban forestry planning and funding for the next fiscal year (Informational Report and Discussion). 5. Review of Urban Forestry Council Prioritized Work Plan for 2008 for selection of one or more items to begin work on and identify action steps to achieve each goal (Explanatory Document: Work Plan Prioritized List for 2008) (Discussion). 6. Staff Report. Staff will provide updates on UFC administrative and programmatic operations relating to research, planning, funding, outreach, and other related activities (Informational Report and Discussion). 7. Committee Reports: (Informational Reports and Discussion). The next meeting is scheduled for June 19, 2008 at 4:15 p.m. at City Hall, Room 421. The next meeting is scheduled for June 10, 2008 at 4:00 p.m. at City Hall, Room 421....
    \end{tcolorbox}
    \caption{An example text excerpt with \textit{Readability} score of 0.}
    \label{fig:read_0}
\end{figure*}

\begin{figure*}
    \centering
    \begin{tcolorbox}[exampleboxstyle, title=Readability Score: 1]
The settings of the story is very desert like what I mean by that is its the desert and its hot and he really had to push his self to make it. The condition at the rode are very rough and has a lot of shape turn it in. Apart in the story he gets a boost of energy and pedal as hard as he can down a hill and then relaxes. This is the setting and is summary of the story…
    \end{tcolorbox}
    \caption{An example text excerpt with \textit{Readability} score of 1.}
    \label{fig:read_1}
\end{figure*}

\begin{figure*}
    \centering
    \begin{tcolorbox}[exampleboxstyle, title=Readability Score: 2]
The features of the setting affect the cyclist because when you have hills to climb and little water, you will get dehydrated. Also the heat from the desert is so hot that it also can make you dehydrated. If you don't pace yourself and don't drink too much water you will be able to reach your goal. Your rest is a big thing for if you don't have energy, you will not get far…
    \end{tcolorbox}
    \caption{An example text excerpt with \textit{Readability} score of 2.}
    \label{fig:read_2}
\end{figure*}

\begin{figure*}
    \centering
    \begin{tcolorbox}[exampleboxstyle, title=Readability Score: 3]
The terrain during the cyclist's journey greatly affects him. For example, the first terrain that he experienced was not very hilly, but rather flat and soothing. The author stated, "I rode into the morning with strong legs and a smile on my face." This shows that he was energized and happy. However, the reader can predict that the journey will not remain this joyful, because the cyclist is basically in the desert during the summer, in which it is extremely hot. Then, the cyclist experiences hilly terrain that sucked the life from his body, especially because he had no water left. The cyclist said, "sometimes life can feel so cruel", emphasizing that the cyclist mood had changed from enthusiastic to tired and forlorn. This change of mood from the terrain can be connected to real life, as obstacles are include, in which the person must persevere and be strong to overcome, in which the cyclist finally did…
    \end{tcolorbox}
    \caption{An example text excerpt with \textit{Readability} score of 3.}
    \label{fig:read_3}
\end{figure*}

\begin{figure*}
    \centering
    \begin{tcolorbox}[exampleboxstyle, title=Readability Score: 4]
People study in college or university for many different reasons. I think the most important reason is to gain more knowledge and learn
more skills. Of course, there are also many other reasons that people study in college such as to get more friends, and increase one's self-confidence. These days, most jobs require people who are educated and have good job skills. Therefore, the people who want a good job have to study hard and at least graduate with a high education. Furthermore, as technology advances allover the world, more and more
education is required of people. Some people who study in college or university want to make more friends and increase their interpersonal skills. They enjoy their lives in university or college and tend to socialize a lot. They can meet more people who have the similar interests with themselves. They can go to uni ball after school and make more friends who they trust. The people who graduate from college seem more confident in our community. These people are more respected by society. Many people want to be respected and to be important by family, friends, their bosses, and others in their lives…
    \end{tcolorbox}
    \caption{An example text excerpt with \textit{Readability} score of 4.}
    \label{fig:read_4}
\end{figure*}

\begin{figure*}
    \centering
    \begin{tcolorbox}[exampleboxstyle, title=Readability Score: 5]
The bar chart and pie chart give information about why US residents travelled and what travel problems they experienced in the year 2009.
It is clear that the principal reason why Americans travelled in 2009 was to commute to and from work. In the same year, the primary
concern of Americans, with regard to the trips they made, was the cost of travelling. Looking more closely at the bar chart, we can see that 49\% of the trips made by Americans in 2009 were for the purpose of commuting. By contrast, only 6\% of trips were visits to friends or relatives, and one in ten trips were for social or recreation reasons. Shopping was cited as the reason for 16\% of all travel, while unspecific 'personal reasons' accounted for the remaining 19\%. According to the pie chart, price was the key consideration for 36\% of American travellers. Almost one in five people cited safety as their foremost travel concern, while aggressive driving and highway congestion were the main issues for 17\% and 14\% of the travelling public. Finally, a total of 14\% of those surveyed thought that access to public transport or space for pedestrians were the most important travel issues...
    \end{tcolorbox}
    \caption{An example text excerpt with \textit{Readability} score of 5.}
    \label{fig:read_5}
\end{figure*}

\begin{figure*}
    \centering
    \begin{tcolorbox}[exampleboxstyle, title=Reasoning Score: 0]
Get answers to the most daunting career questions here! Are you looking for a speaker for an upcoming conference? Look no further! I would love to connect with you on social media or an upcoming event! Top 5 Things to Consider BEFORE you Quit! Join my email list and receive the missing pieces to your successful career right in your inbox!
    \end{tcolorbox}
    \caption{An example text excerpt with \textit{Reasoning} score of 0.}
    \label{fig:reas_0}
\end{figure*}

\begin{figure*}
    \centering
    \begin{tcolorbox}[exampleboxstyle, title=Reasoning Score: 1]
Light is the source of life on earth. Take the advantage of the sunlight by guiding it into any of your rooms with The Viva glass doors collection. This glass door generates a bright and friendly atmosphere and explores a new sense of space. The Viva internal glass door range is created around innovative engineering, quality workmanship and attractive design. Due to its minimalist style of a crystal clear surface with a frosted design, the Viva glass door collection integrates harmoniously into any room. The aesthetics of modern home decor is characterized by simple and vibrant elegance. With Viva internal glass double doors, lightness and transparency is generated by their crystal clear surfaces with minimalist frosted designs. The Viva glass door collection, to meet the bespoke requirements, can be manufactured in sizes up to (w)1600mm X (h)2500mm.
    \end{tcolorbox}
    \caption{An example text excerpt with \textit{Reasoning} score of 1.}
    \label{fig:reas_1}
\end{figure*}

\begin{figure*}
    \centering
    \begin{tcolorbox}[exampleboxstyle, title=Reasoning Score: 2]
Deep in the eastern base of the Whetstone Mountains – Southern Arizona, sit the most protected water-filled caves in America. Keeping them natural is no easy task. A solution to conserving thousands of gallons of cave water has been discovered. 'Saving the Caves' is another great story of how collecting rainwater can pay for itself. In this video, the water collected off of a small building roof is used to preserve the natural water in the caves located at the Kartchner Caverns State Park in Arizona. This state park is home to the last natural water-filled caves in America. To learn about how these caves were discovered, you can read their story here. If you have a story about how you used rainwater to improve your living conditions, I would love to hear it. Comment below and I will contact you to learn more.
    \end{tcolorbox}
    \caption{An example text excerpt with \textit{Reasoning} score of 2.}
    \label{fig:reas_2}
\end{figure*}

\begin{figure*}
    \centering
    \begin{tcolorbox}[exampleboxstyle, title=Reasoning Score: 3]
Hull Venue, a new 3,500-seat, multi-purpose complex in the English city, has revealed details of its first four shows ahead of its opening later this year. Cult comedy The League of Gentlemen will visit Hull Venue on September 4. The show was co-created by, and stars, Hull's Reece Shearsmith. The event will form part of the show's first UK tour in more than 12 years. The venue will also play host to a 'Strictly Come Dancing – The Professionals' event in the spring of 2019. The event will feature some of the BBC programme's professional dancers and has been pencilled in for May 19, 2019...
    \end{tcolorbox}
    \caption{An example text excerpt with \textit{Reasoning} score of 3.}
    \label{fig:reas_3}
\end{figure*}

\begin{figure*}
    \centering
    \begin{tcolorbox}[exampleboxstyle, title=Reasoning Score: 4]
This is devastating. Lindsey Marie Michaels, a 21-year-old perfusion student at Carlow University, died in Pittsburgh, PA, after train hopping with her boyfriend. The young man, who has not been identified, only sustained an ankle injury. According to Urban Dictionary, train hopping is "a term used when using a subway and walking from one subway to another at the arrival of a station. Common uses of train hopping are when your exit at the station is at a certain place and you want to get as close to it as possible when the Subway comes to a stop at your station." The incident took place near South Eighth Street at roughly 2:30 a.m. on Sunday. The train continued along the Norfolk Southern train tracks and stopped in Etna about 25 minutes later, according to the Pittsburgh Post-Gazette, "after being alerted by Pittsburgh authorities about a possible pedestrian fatality involving the train." This activity is extremely dangerous, and illegal, leading to jail time or a hefty fine in some states. According to the MTA, in NYC, violators could be forced to pay a \$100 fine for both fare evasion and interference with movement. Lindsey\'s school made a statement on behalf of the tragedy via the publication which said...
    \end{tcolorbox}
    \caption{An example text excerpt with \textit{Reasoning} score of 4.}
    \label{fig:reas_4}
\end{figure*}

\begin{figure*}
    \centering
    \begin{tcolorbox}[exampleboxstyle, title=Reasoning Score: 5]
Having built a reputation as an exceptional reedman in Seattle, Dave Anderson presents a sparkling debut on the melodically rich Clarity, alternating between alto and soprano saxophones on eight original compositions and two covers. Having performed extensively throughout North America with luminaries like Jim McNeely, Clark Terry and the late great Mel Torme, Anderson moved to Seattle in 2005 from his native Minnesota, forming Dave Anderson Quartet after a one-nighter at Egan\'s Ballard Jamhouse. The group consists of pianist John Hansen; bassist Chuck Kistler; and drummer Adam Kessler, with Thomas Marriott taking to the flugelghorn in a guest appearance on "Wabi-Sabi." Anderson\'s compositions are impressive, offering a varied selection of tones and harmonies, though he chooses to open the set with Joe Henderson\'s spicy samba, "Y Ya La Quiero," exploring it with his soprano voice, masterfully accompanied by Hansen. The frontline duet of Marriott and Anderson (again on soprano) on "Wabi-Sabi," is something sweet and special, while "Stalemate" is the first tune to display Anderson\'s alto chops, and presents Kistler\'s first solo...
    \end{tcolorbox}
    \caption{An example text excerpt with \textit{Reasoning} score of 5.}
    \label{fig:reas_5}
\end{figure*}

\begin{figure*}
    \centering
    \begin{tcolorbox}[exampleboxstyle, title=Cleanliness Score: 0]
"Somewhere Over the Rainbow" https://www.youtube.com/watch?v=h7AV1jQple4 "Part of Your World" https://www.youtube.com/watch?v=pUlit0d3Uu8
"Falling Slowly" (from Once) https://www.youtube.com/watch?v=VkkD3xtpTiw "Vanilla Ice Cream" https://www.youtube.com/watch?v=F2gLraxpBEE
    \end{tcolorbox}
    \caption{An example text excerpt with \textit{Cleanliness} score of 0.}
    \label{fig:clean_0}
\end{figure*}

\begin{figure*}
    \centering
    \begin{tcolorbox}[exampleboxstyle, title=Cleanliness Score: 1]
adget will be
* reloaded from scratch. This function will be passed one
parameter, an
* opensocial.ResponseItem. The error code will be set to reflect
whether
*
there were any problems with the request. If there was no error, the
* message was sent. If there was an error, you can use the
response item's
* getErrorCode method
to determine how to proceed. The data on the response
* item will not be set.
*
* @member opensocial
*
@private
*/
opensocial.Container.prototype.requestSendMessage =
function(recipients,
    \end{tcolorbox}
    \caption{An example text excerpt with \textit{Cleanliness} score of 1.}
    \label{fig:clean_1}
\end{figure*}

\begin{figure*}
    \centering
    \begin{tcolorbox}[exampleboxstyle, title=Cleanliness Score: 2]
Designs that will not lose their sense of unity can be seen where the fluffy and tender impression is tightened in black.
At Google, which attracts people with various functions, the office has the charm of each branch.
Among them, the Swiss branch office is an office where there is a sense of unity that there is no sense of unity.
There are unique areas ranging from egg-shaped private rooms to rooms simulating grasslands and garages, ski resorts and kamakura types.
The various chairs symbolize the entire office.
Cybozu Inc. A giraffe welcomes you at the entrance.
In the office meeting space, the table is a whiteboard, A variety of ingenuity has been applied, such as a uniquely
shaped sofa. London 's advertising agency, Mother London, is a simple, modern concrete-coated room.    
\end{tcolorbox}
    \caption{An example text excerpt with \textit{Cleanliness} score of 2.}
    \label{fig:clean_2}
\end{figure*}

\begin{figure*}
    \centering
    \begin{tcolorbox}[exampleboxstyle, title=Cleanliness Score: 3]
Strange Country Day by Charles Curtis!! "You\'re like all the tourists. You can\'t stop looking at all the pretty lights," she said as we weaved our way through the foot traffic. "I guess." We stopped at a corner and waited for a green light. Sophi looked down the block at the scene, the endless colored lights dancing on her face. She stared up at one of the signs, which featured a massive cup of soup with actual steam rising out of it. "I can sort of see it. Just imagine how much energy it takes to keep everything running," she said as she put her hand against the streetlamp. The light above her immediately went out, as did the stoplight connected to it. I opened my mouth to wonder aloud what had happened, but nothing came out as we watched as the signs ahead of us began shutting down one by one. One second a screen was filled with a skinny woman drinking a soda...
    \end{tcolorbox}
    \caption{An example text excerpt with \textit{Cleanliness} score of 3.}
    \label{fig:clean_3}
\end{figure*}

\begin{figure*}
    \centering
    \begin{tcolorbox}[exampleboxstyle, title=Cleanliness Score: 4]
Locally owned and operated, Iowa Running Company is your premier run specialty shop for Cedar Rapids and the greater Corridor. Don\'t let the name fool you. We are more than a shop for "just runners". With a shopping atmosphere and experience like no other, we\'re sure you\'ll never want to leave! But don\'t take our word for it. Come check us out next time you\'re walking around the NewBo District! With over 20 years of run specialty experience nationally and locally, we have some of the best shoe fitters around. Our friendly and knowledgeable staff will always be upfront and honest with you, and help guide you to make the most educated decision that best suites your active endeavors...
    \end{tcolorbox}
    \caption{An example text excerpt with \textit{Cleanliness} score of 4.}
    \label{fig:clean_4}
\end{figure*}

\begin{figure*}
    \centering
    \begin{tcolorbox}[exampleboxstyle, title=Cleanliness Score: 5]
Macro Media Lab Stay on top of trends in emerging markets Want to find the next Epic Games? You should be looking at India. Post author By Daniel Tuba D\'Souza No Comments on Want to find the next Epic Games? You should be looking at India. The esports and mobile gaming industries in India is one of the fastest growing markets in the world. It\'s on track to surpass than the music, movie, and television industries put together. Hello, I\'m Daniel and welcome to Macro Media Lab! Twice a month I tackle some of the largest trends in emerging markets and break them down so you can understand how the world is changing. This weeks report is on the e-sports industry in India. The Quick Summary The esports industry in India is expected to 3.7x over the next 4 years. Growing from \$935M USD to \$3.7B USD...
    \end{tcolorbox}
    \caption{An example text excerpt with \textit{Cleanliness} score of 5.}
    \label{fig:clean_5}
\end{figure*}

\section{Full Experimental Results}
\label{app:results}

\subsection{Main Experiment}
\label{app:dataselection}
The full results of data selection methods for pre-training 1.3B model are shown in Table \ref{tab:fullselection}.

\begin{table*}[!tb]
\centering
\begin{tabular}{@{}lcccccccc@{}}
\toprule
\textbf{Data Selection Method} & \textbf{ARC-E} & \textbf{ARC-C} & \textbf{SciQ} & \textbf{SIQA} & \textbf{WG} & \textbf{HS} & \textbf{RACE} & \textbf{OBQA} \\ \midrule
Random (30B tokens)                & 51.05 & 23.81 & 83.50 & 40.28 & 51.85 & 39.69 & 30.43 & 29.60 \\
Random (60B tokens)                & 54.25 & 26.79 & 87.00 & 39.97 & 53.20 & 41.45 & 30.53 & 32.40 \\ \midrule
PPL                                & 49.71 & 25.09 & 82.80 & 37.72 & 49.80 & 34.06 & 25.84 & 27.20 \\ \midrule
Semdedup                           & 50.59 & 24.66 & 82.70 & 38.89 & 50.67 & 38.41 & 30.43 & 27.40 \\ \midrule
DSIR                               &       &       &       &       &       &       &       &       \\
\quad Target as \textit{Book}      & 49.49 & 24.57 & 83.30 & 42.48 & 54.38 & 43.92 & 24.88 & 33.00 \\
\quad Target as \textit{Wikipedia} & 54.34 & 26.19 & 84.30 & 38.28 & 51.78 & 35.55 & 24.78 & 30.00 \\ \midrule
QuRating                           &       &       &       &       &       &       &       &       \\
\quad \textit{Required Expertise}  & 58.27 & 28.86 & 83.60 & 39.92 & 53.12 & 42.44 & 24.11 & 32.00 \\
\quad \textit{Writing Style}       & 57.58 & 28.24 & 85.60 & 41.15 & 53.83 & 43.85 & 24.98 & 31.40 \\
\quad \textit{Facts and Trivia}    & 58.96 & 29.27 & 84.50 & 40.58 & 53.12 & 43.16 & 26.60 & 32.20 \\
\quad \textit{Educational Value}   & 58.73 & 29.94 & 84.30 & 41.35 & 54.14 & 44.66 & 24.59 & 31.60 \\ \midrule
Fineweb-Edu                        & 55.13 & 28.14 & 84.10 & 41.71 & 53.91 & 40.90 & 31.20 & 31.00 \\ \midrule
MATES                              & 52.60 & 24.25 & 82.60 & 38.69 & 52.17 & 38.90 & 32.10 & 29.00 \\ \midrule
\textbf{PRRC (Ours)}               &       &       &       &       &       &       &       &       \\
\quad \textit{Professionalism}     & 55.85 & 27.56 & 84.92 & 39.99 & 52.78 & 41.20 & 29.98 & 29.80 \\
\quad \textit{Readability}         & 55.64 & 26.19 & 86.70 & 40.17 & 53.16 & 42.89 & 32.00 & 30.40 \\
\quad \textit{Reasoning}           & 55.35 & 27.05 & 84.30 & 40.36 & 52.87 & 41.34 & 30.95 & 30.00 \\
\quad \textit{Cleanliness}         & 56.89 & 27.65 & 84.80 & 41.97 & 52.33 & 40.34 & 30.24 & 31.20 \\ \midrule
\textbf{Meta-rater (Ours)}         &       &       &       &       &       &       &       &       \\
\quad PRRC (4)                     & 56.87 & 28.16 & 86.00 & 42.28 & 52.67 & 42.63 & 30.62 & 31.60 \\
\quad Model (11)                   & 56.48 & 28.75 & 86.80 & 43.05 & 53.85 & 39.97 & 31.72 & 32.20 \\
\quad All (25)                     & 58.25 & 29.86 & 88.60 & 42.68 & 53.75 & 39.81 & 31.10 & 32.00 \\ \bottomrule
\end{tabular}
\caption{Full downstream tasks results of data selection methods. Abbreviations: WG = WinoGrande,  HS = HellaSwag, OBQA = OpenbookQA.}
\label{tab:fullselection}
\end{table*}

\subsection{Scaling Experiment}
\label{app:scaling}
The full results of scaling experiment are provided in Table \ref{tab:fullscaling}.
Moreover, we also conducted scaling law experiments on smaller models (178M and 407M), with results shown in Table \ref{tab:smaller}.

\begin{table*}
\centering
\begin{tabular}{@{}clllll@{}}
\toprule
\textbf{Model Size} & \textbf{Method}        & \textbf{G.K.}  & \textbf{C.R.}  & \textbf{R.C.}  & \textbf{Avg.} \\ \midrule
\multirow{3}{*}{178M} & Random                     & 32.65 & 36.74 & 22.31          & 31.60 \\
                      & Qurating-\textit{Educational Value} & 32.79 & 36.60 & 22.31          & 31.60 \\
                    & \textbf{Meta-rater (Ours)} & \textbf{33.00} & \textbf{36.97} & \textbf{23.26} & \textbf{32.05}   \\ \midrule
\multirow{3}{*}{407M} & Random                     & 39.05 & 37.68 & 23.66          & 34.69 \\
                      & Qurating-\textit{Educational Value} & 41.96 & 37.69 & \textbf{25.74} & 36.30 \\
                    & \textbf{Meta-rater (Ours)} & \textbf{42.15} & \textbf{37.72} & 25.67          & \textbf{36.37}   \\ \bottomrule
\end{tabular}
\caption{Downstream tasks results of smaller models.}
\label{tab:smaller}
\end{table*}

\subsection{Combination Strategy Experiment}
\label{app:combination_strategy}
The full results of combination strategy experiment are provided in Table \ref{tab:fullcomb}.

\subsection{Analysis of Proxy Models}
\label{app:regressionrun}
The full results of proxy model analysis experiment are shown in Table \ref{tab:fullreg}.
\begin{table*}
\centering
\begin{tabular}{@{}cccccccccc@{}}
\toprule
\multicolumn{1}{l}{\textbf{Model Size}} &
  \textbf{Method} &
  \textbf{ARC-E} &
  \textbf{ARC-C} &
  \textbf{SciQ} &
  \textbf{SIQA} &
  \textbf{WG} &
  \textbf{HS} &
  \textbf{RACE} &
  \textbf{OBQA} \\ \midrule
\multirow{2}{*}{3.3B} & Random     & 66.33 & 33.53 & 92.80 & 43.71 & 59.59 & 57.35 & 34.35 & 36.20 \\
                      & Meta-rater & 72.10 & 37.54 & 92.90 & 43.91 & 60.14 & 58.99 & 35.12 & 37.00 \\ \midrule
\multirow{2}{*}{7.2B} & Random     & 67.77 & 36.43 & 91.10 & 42.73 & 60.29 & 53.02 & 34.73 & 37.00 \\
                      & Meta-rater & 71.34 & 39.76 & 92.80 & 44.32 & 60.45 & 58.97 & 36.08 & 38.20 \\ \bottomrule
\end{tabular}
\caption{Full downstream tasks results of 3.3B and 7.2B models.}
\label{tab:fullscaling}
\end{table*}

\begin{table*}
\centering
\begin{tabular}{@{}lcccccccc@{}}
\toprule
\textbf{Combination Strategy} & \textbf{ARC-E} & \textbf{ARC-C} & \textbf{SciQ} & \textbf{SIQA} & \textbf{WG} & \textbf{HS} & \textbf{RACE} & \textbf{OBQA} \\ \midrule
Mean               &       &       &       &       &       &       &       &       \\
\quad PRRC (4)     & 53.91 & 26.62 & 84.60 & 38.13 & 52.09 & 37.04 & 29.67 & 31.00 \\
\quad Model (11)   & 58.12 & 28.92 & 87.10 & 39.41 & 50.28 & 37.46 & 31.39 & 31.20 \\
\quad All (25)     & 56.99 & 28.67 & 83.20 & 37.97 & 51.70 & 37.22 & 31.48 & 30.00 \\ \midrule
Intersection       &       &       &       &       &       &       &       &       \\
\quad QuRating (4) & 52.64 & 27.37 & 83.70 & 39.95 & 51.30 & 36.96 & 31.39 & 31.20 \\
\quad PRRC (4)     & 53.67 & 27.81 & 86.00 & 40.28 & 52.17 & 39.70 & 30.72 & 31.00 \\ \bottomrule
\end{tabular}
\caption{Full downstream tasks results of combination strategy experiment.}
\label{tab:fullcomb}
\end{table*}

\begin{table*}
\centering
\begin{tabular}{@{}ccccccccc@{}}
\toprule
\multicolumn{1}{l}{\textbf{$N$}} & \textbf{ARC-E} & \textbf{ARC-C} & \textbf{SciQ} & \textbf{SIQA} & \textbf{WG} & \textbf{HS} & \textbf{RACE} & \textbf{OBQA} \\ \midrule
128 & 57.31 & 26.96 & 84.60 & 40.92 & 53.12 & 42.66 & 30.00 & 31.80 \\
256 & 58.25 & 29.86 & 88.60 & 42.68 & 53.75 & 39.81 & 31.10 & 32.00 \\
512 & 59.71 & 29.86 & 88.00 & 42.68 & 54.54 & 40.44 & 31.00 & 32.00 \\ \bottomrule
\end{tabular}
\caption{Full downstream tasks results of proxy model analysis experiment.}
\label{tab:fullreg}
\end{table*}

\begin{table*}
\centering
\begin{tabular}{@{}ccccccccc@{}}
\toprule
\multicolumn{1}{l}{\textbf{Data Domain}} & \textbf{ARC-E} & \textbf{ARC-C} & \textbf{SciQ} & \textbf{SIQA} & \textbf{WG} & \textbf{HS} & \textbf{RACE} & \textbf{OBQA} \\ \midrule
All Domains                              & 58.25          & 29.86          & 88.60         & 42.68         & 53.75       & 39.81       & 31.10         & 32.00         \\
CC-Only                                  & 55.68          & 28.84          & 86.60         & 41.02         & 50.83       & 36.05       & 31.10         & 31.60         \\ \bottomrule
\end{tabular}
\caption{Full downstream tasks results of data domain analysis experiment.}
\label{tab:fulldomain}
\end{table*}

\subsection{Analysis of Data Domain}
\label{app:data_domain}
The full results of data domain analysis experiment are shown in Table \ref{tab:fulldomain}.

\section{Evalution Results on MMLU and NaturalQuestions}
\label{app:MMLU_NQ_results}
We also evaluate pre-trained models on two challenging tasks, with results shown in Table \ref{tab:MMLU_NQ}. 
Our analysis reveals several important insights:

1. Scale limitations: For MMLU, all 1.3B models perform near random-chance level (25\%), confirming previous findings that smaller models struggle with this benchmark. This aligns with observations in prior work \cite{regmix,qurating} showing that models below 7B parameters typically perform at or slightly above random chance on MMLU regardless of training methodology.

2. Consistent patterns: Despite the overall low performance, Meta-rater still shows a slight improvement over random selection in NaturalQuestions for both model scales (2.30\% vs. 2.13\% for 1.3B; 6.87\% vs. 6.28\% for 3.3B). This suggests our method's benefits extend to knowledge-intensive tasks, though the absolute performance remains limited by model capacity.

3. Scaling effects: The significant jump in NaturalQuestions performance from 1.3B to 7.2B models (approximately 5x improvement) indicates that model scale is particularly important for knowledge-intensive tasks. This is consistent with the literature showing that knowledge retrieval capabilities improve non-linearly with model size.

\begin{table*}
\centering
\begin{tabular}{@{}clcc@{}}
\toprule
\multicolumn{1}{l}{\textbf{Model}} & \textbf{Method} & \multicolumn{1}{l}{\textbf{MMLU}} & \multicolumn{1}{l}{\textbf{NQ}} \\ \midrule
\multirow{3}{*}{1.3B} & Random                     & 25.99 & 2.13  \\
                      & QuRating-\textit{Educational Value} & 26.73 & 2.05  \\
                      & \textbf{Meta-rater}        & 25.89 & 2.30  \\ \midrule
\multirow{2}{*}{3.3B} & Random                     & 25.48 & 6.28  \\
                      & \textbf{Meta-rater}        & 26.21 & 6.87  \\ \midrule
\multirow{2}{*}{7.2B} & Random                     & 26.21 & 10.89 \\
                      & \textbf{Meta-rater}        & 26.24 & 10.42 \\ \bottomrule
\end{tabular}
\caption{Downstream tasks results on challenging tasks.}
\label{tab:MMLU_NQ}
\end{table*}

\end{document}